\documentclass[10pt,twocolumn,letterpaper]{article}

\usepackage{wacv}              

\definecolor{wacvblue}{rgb}{0.21,0.49,0.74}
\usepackage[pagebackref,breaklinks,colorlinks,allcolors=wacvblue]{hyperref}

\def\wacvPaperID{26} 
\def\confName{WACV}
\def\confYear{2026}

\title{ViTNT-FIQA: Training-Free Face Image Quality Assessment with Vision Transformers}

\author{
Guray Ozgur$^{1,2}$,
Eduarda Caldeira$^{1,2}$,
Tahar Chettaoui$^{1,2}$,
Jan Niklas Kolf$^{1,2}$,\\
Marco Huber$^{1,2}$,
Naser Damer$^{1,2}$,
Fadi Boutros$^{1}$ \\
$^{1}$Fraunhofer IGD, Germany,
$^{2}$TU Darmstadt, Germany}

\usepackage{graphics} 
\usepackage{amsmath} 
\usepackage{amssymb}  
\usepackage{booktabs}
\usepackage{multirow}
\usepackage{rotating}
\usepackage{tikz}
\usepackage{pifont}
\usepackage{enumitem}

\newcommand{\cmark}{\ding{51}}%
\newcommand{\xmark}{\ding{55}}%
\usepackage{xspace}
\newcommand{\grafiqs}{\textsc{GraFIQs}\xspace}
\newcommand{\ourmethod}{\textit{ViTNT-FIQA}\xspace}

\begin{document}

\maketitle

\begin{abstract}
Face Image Quality Assessment (FIQA) is essential for reliable face recognition systems. Current approaches primarily exploit only final-layer representations, while training-free methods require multiple forward passes or backpropagation. We propose ViTNT-FIQA\footnote{The implementation is publicly available at: \url{https://github.com/gurayozgur/ViTNT-FIQA}}, a training-free approach that measures the stability of patch embedding evolution across intermediate Vision Transformer (ViT) blocks. We demonstrate that high-quality face images exhibit stable feature refinement trajectories across blocks, while degraded images show erratic transformations. Our method computes Euclidean distances between L2-normalized patch embeddings from consecutive transformer blocks and aggregates them into image-level quality scores. We empirically validate this correlation on a quality-labeled synthetic dataset with controlled degradation levels. Unlike existing training-free approaches, ViTNT-FIQA requires only a single forward pass without backpropagation or architectural modifications. Through extensive evaluation on eight benchmarks (LFW, AgeDB-30, CFP-FP, CALFW, Adience, CPLFW, XQLFW, IJB-C), we show that ViTNT-FIQA achieves competitive performance with state-of-the-art methods while maintaining computational efficiency and immediate applicability to any pre-trained ViT-based face recognition model.
\vspace{-3mm}
\end{abstract}

\begin{figure}[t]
    \centering
    \includegraphics[width=0.45\textwidth]{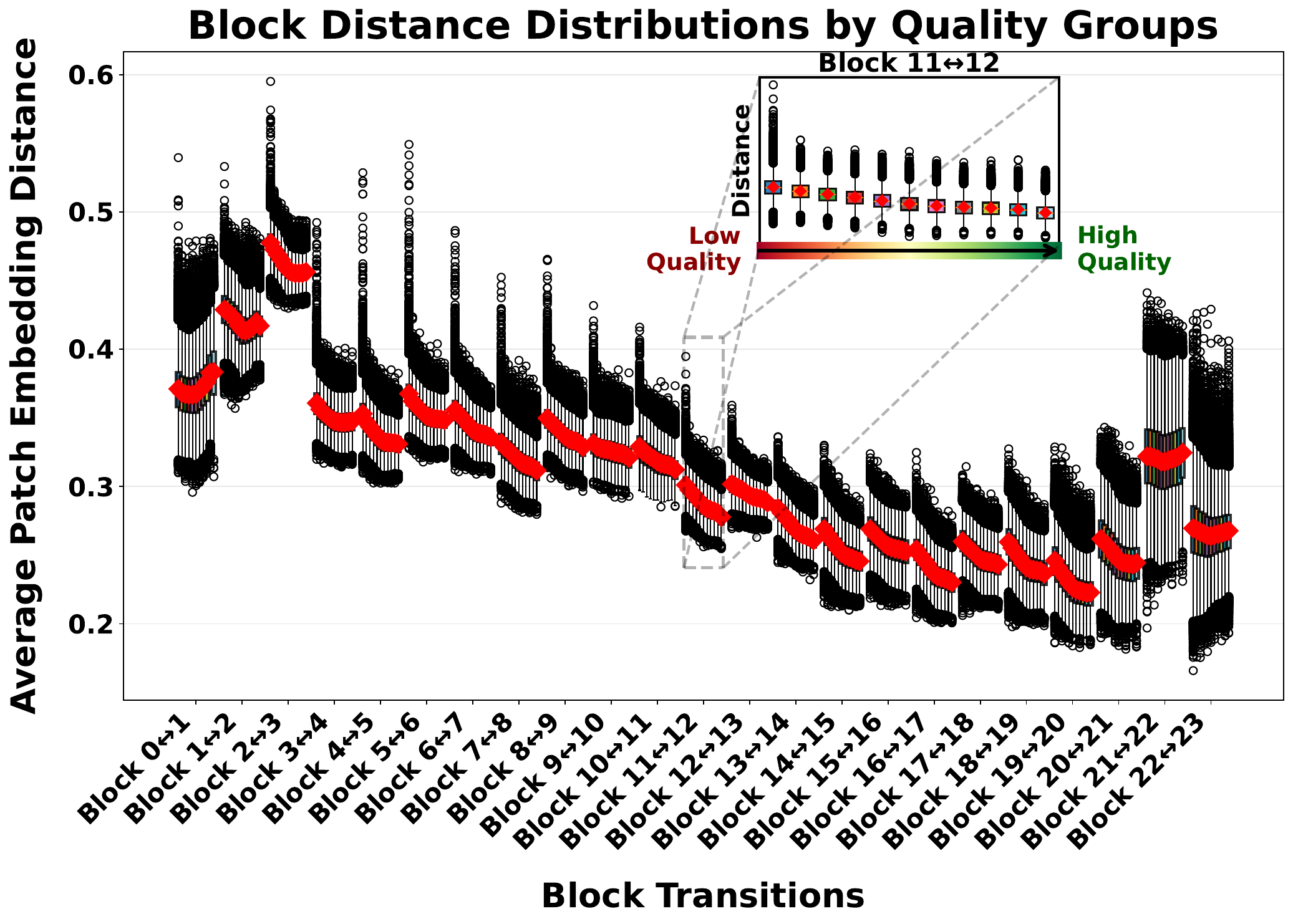}
    \caption{Boxplots of mean L2 distances between corresponding patch embeddings from consecutive ViT-B blocks computed for 11 quality groups, each having 0.5M images, from 5.5M images of SynFIQA \cite{mrfiqa}. Each box summarizes the distribution of average patch-embedding distances across images in a quality group, lower distances empirically correspond to higher ground-truth quality for most block transitions. The inset (Block 11 $\leftrightarrow$ 12) shows the quality gradient (low $\rightarrow$ high) and illustrates how the distances across groups provide a measure of quality discriminability, i.e. the higher the quality, the lower the distance.}
    \label{fig:synfiqa}
    \vspace{-6mm}
\end{figure}

\begin{figure*}[ht]
\centering
\includegraphics[width=0.83\textwidth, trim=1 1 1 1,clip]{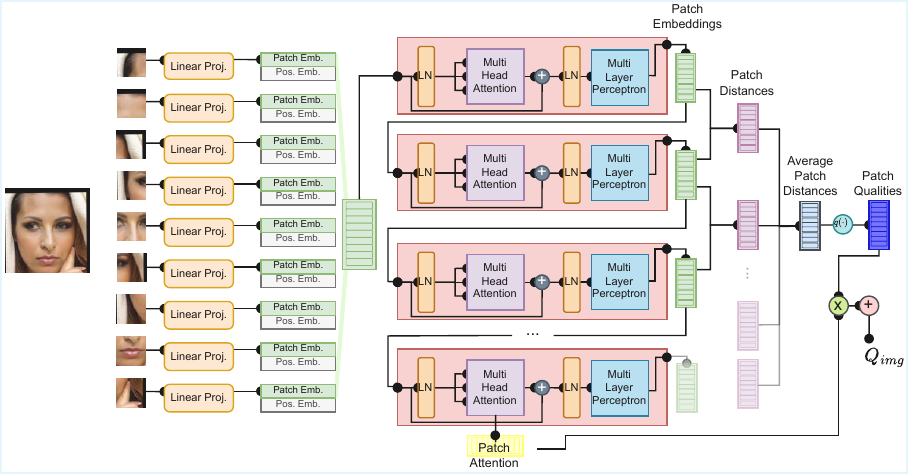}
\caption{Overview of our ViT-based quality assessment method \ourmethod. (1) The face image is patchified and embedded. (2) Intermediate patch representations are extracted from selected transformer blocks. (3) L2-normalized embeddings are compared across consecutive blocks to measure patch-level feature distances. (4) Distances are mapped to quality scores per patch level, which are aggregated, uniformly or using attention weights, to produce the final image-level quality estimate.}
\label{fig:pipeline}
\vspace{-6mm}
\end{figure*}

\vspace{-2mm}
\section{Introduction}
\label{sec:intro}

FIQA evaluates the utility of face images for face recognition (FR), specifically measuring \textit{recognition utility} or \textit{suitability for identity verification} \cite{Quality_ISO,NISTQuaity}. Unlike general Image Quality Assessment (IQA) methods that assess quality from human perception \cite{BRISQE_IQA,nique,liu2017rankiqa}, FIQA quantifies how effectively a facial image serves automated recognition tasks. As demonstrated in \cite{BiyingWACV}, high perceived quality does not always correlate with FR utility, particularly when factors like facial occlusions are present. Current FIQA approaches primarily exploit only final-layer representations from deep networks \cite{PFE_FIQA,SERFIQ,MagFace,boutros_2023_crfiqa,grafiqs,atzori2025vitfiqaassessingfaceimage}. Training-free methods, while offering immediate applicability to pre-trained models, typically require either multiple forward passes with varied dropout patterns \cite{SERFIQ} or backpropagation \cite{grafiqs,FaceQAN}, increasing computational overhead (Table \ref{tab:difference_methods}). Recent research on ViT internals has revealed that transformer blocks refine features iteratively with high inter-block similarity \cite{raghu2021vit}, where residual connections propagate information forward and each block produces slight refinements. This smooth feature evolution trajectory suggests that the stability of patch representations across intermediate blocks may contain quality-relevant information, yet this remains unexplored for FIQA. We propose a \textit{\textbf{ViT}-based \textbf{N}o-\textbf{T}raining \textbf{FIQA}} approach, hence the name \textbf{\ourmethod},  which analyzes the stability of patch embedding evolution across intermediate transformer blocks in pre-trained ViT-based models. Our method is grounded in the hypothesis that high-quality face images exhibit smoother, more stable feature refinement trajectories across blocks, while degraded images show erratic transformations. We empirically validate this hypothesis on SynFIQA \cite{mrfiqa}, a quality-labeled synthetic dataset with 550,000 images across controlled degradation levels (Figure \ref{fig:synfiqa}), demonstrating that cross-block patch embedding distances systematically decrease with increasing ground-truth quality across most transformer block transitions. Unlike existing approaches, our \ourmethod does not make use of any quality labels \cite{faceqnetv1,SDDFIQA}, any training \cite{PFE_FIQA,boutros_2023_crfiqa}, or any custom loss \cite{MagFace}. Moreover, different from training-free approaches \cite{SERFIQ,grafiqs}, it only requires a single forward pass without backpropagation. Clear conceptual comparisons to the state-of-the-art (SOTA) methods are shown in Table \ref{tab:difference_methods}. We make the following contributions:
\begin{itemize}
\vspace{-1.1mm}
    \item A training-free FIQA method that measures patch-level cross-block distances in pre-trained ViT models, requiring only a single forward pass without backpropagation or architectural modifications.
    \item A comprehensive analysis demonstrating that cross-block embedding stability correlates with face image quality, providing a novel quality indicator. 
    \item Extensive evaluation across eight benchmark datasets (LFW\cite{LFWTech}, AgeDB-30\cite{agedb}, CFP-FP\cite{cfp-fp}, CALFW\cite{CALFW}, Adience\cite{Adience}, CPLFW\cite{CPLFWTech}, XQLFW\cite{XQLFW}, IJB-C\cite{ijbc}) demonstrating competitive performance with existing SOTA methods.
\end{itemize}

\begin{table}[t]
\centering
\caption{Conceptual comparison on the design choices between our \ourmethod and recent FIQA approaches in the literature.}
\label{tab:difference_methods}
\vspace{-2mm}
\resizebox{0.95\linewidth}{!}{%
\begin{tabular}{cccccccc}
\multicolumn{1}{l}{} & \multicolumn{1}{l}{} & \multicolumn{1}{l}{} & \multicolumn{1}{l}{} & \multicolumn{4}{c}{Inference} \\
\cline{5-8} \\
\multicolumn{1}{c}{\begin{sideways}FIQA\end{sideways}} & \multicolumn{1}{l}{\begin{sideways}\begin{tabular}[c]{@{}l@{}}Quality \\Labels\end{tabular}\end{sideways}} &   \multicolumn{1}{l}{\begin{sideways}\begin{tabular}[c]{@{}l@{}}Requires \\Training\end{tabular}\end{sideways}} & \multicolumn{1}{l}{\begin{sideways}\begin{tabular}[c]{@{}l@{}}Custom\\ Loss\end{tabular}\end{sideways}} & \multicolumn{1}{l}{\begin{sideways}\begin{tabular}[c]{@{}l@{}}Feed-\\Forwards\end{tabular}\end{sideways}} & \multicolumn{1}{l}{\begin{sideways}Backwards\end{sideways}}  \\
\hline
PFE \cite{PFE_FIQA} & \xmark &   \cmark & \cmark & 1 & 0 \\
SER-FIQ \cite{SERFIQ} & \xmark &   \xmark & \xmark & 100 & 0 \\
FaceQnet \cite{faceqnetv1} & \cmark  & \cmark & \xmark & 1 & 0 \\
MagFace \cite{MagFace} & \xmark &   \cmark & \cmark & 1 & 0  \\
SDD-FIQA \cite{SDDFIQA} & \cmark &  \cmark & \xmark & 1 & 0 \\
CR-FIQA \cite{boutros_2023_crfiqa} & \xmark   & \cmark & \cmark & 1 & 0\\
DifFIQA \cite{10449044} & \xmark  & \cmark & \cmark & 1 & 0\\
eDifFIQA \cite{babnikTBIOM2024} & \cmark  & \cmark & \cmark & 1 & 0  \\
\grafiqs \cite{grafiqs} & \xmark  & \xmark & \xmark & 1 & 1 \\
CLIB-FIQA \cite{Ou_2024_CVPR} & \xmark   & \cmark & \cmark & 1 & 0\\
VIT-FIQA \cite{atzori2025vitfiqaassessingfaceimage} & \xmark   & \cmark & \xmark & 1 & 0\\
\hline
\ourmethod (Ours) & \xmark & \xmark & \xmark & 1 & 0   \\
\end{tabular}
}\vspace{-6mm}
\end{table}

\section{Related Work}
\label{sec:related}

\subsection{Vision Transformer Internals}
\label{subsec:vit_blocks}

ViTs \cite{DBLP:conf/iclr/DosovitskiyB0WZ21} have been successfully applied to FR \cite{Zhong_2021_FT,Dan_2023_TransFace,DBLP:conf/bmvc/SunT22,Kim_2023_SViT,DBLP:conf/cvpr/KimS0JL24,DBLP:journals/ivc/ChettaouiDB25} and recently to FIQA \cite{atzori2025vitfiqaassessingfaceimage}, demonstrating their effectiveness in modeling facial features. Unlike CNNs that process images through hierarchical local operations with gradually expanding receptive fields \cite{lenet5}, ViTs divide images into patches and model global relationships through self-attention mechanisms \cite{DBLP:conf/nips/VaswaniSPUJGKP17}, enabling long-range dependency modeling from the first layer onward.

\textbf{Feature Refinement and Representation Similarity:} Research on ViT internals has revealed that transformer blocks refine features iteratively. Raghu et al. \cite{raghu2021vit} demonstrated through centered kernel alignment (CKA) analysis that ViTs maintain highly similar representations across all layers, exhibiting "much more uniform representations" compared to CNNs which show distinct stage boundaries. Their layer-wise similarity heatmaps revealed a solid grid pattern in ViTs, contrasting with the clear low/high stage gaps observed in ResNets. This uniform similarity structure indicates that each ViT block refines features incrementally, with cosine similarity between successive blocks remaining high throughout the network.

\textbf{Role of Residual Connections:} Skip connections in ViTs are found to be even more influential than in ResNets, having strong effects on performance and representation similarity by removing skip connections, which causes representations before and after that block to become dissimilar and results in accuracy degradation  \cite{raghu2021vit}. This demonstrates that much of the feature information is carried forward via identity paths, with each block's output being a slight refinement of its input rather than a complete transformation. The global self-attention mechanism aggregates context early while residual connections propagate low-level features forward, ensuring gradual enhancement across blocks \cite{raghu2021vit}.

\textbf{Leveraging Intermediate Representations:} The understanding of ViT feature evolution has motivated research on utilizing intermediate representations, specifically on early exits \cite{DBLP:conf/bmvc/BakhtiarniaZI21, slvit, lgvit, distillation_multiexit, BERxiT}. Early exit mechanisms \cite{DBLP:conf/bmvc/BakhtiarniaZI21, slvit, lgvit, distillation_multiexit, BERxiT} allow inference to terminate at intermediate blocks, exploiting the fact that different depths capture distinct levels of feature abstraction. The effectiveness of early exits demonstrates that intermediate block representations contain valuable information beyond serving as stepping stones to final outputs \cite{eesurvey3, eesurvey4}.

Our \ourmethod is also motivated by these insights into ViT's smooth feature refinement trajectory. We empirically validate that the magnitude of change between consecutive blocks, reflecting the degree of feature transformation, can reveal quality-relevant information about face images, see Figure \ref{fig:synfiqa}. Given that ViTs refine features gradually with high inter-block similarity \cite{raghu2021vit}, we investigate whether analyzing the stability of patch embedding evolution across multiple transformer blocks can distinguish high-utility samples from low-utility samples.

\subsection{Face Image Quality Assessment}
\label{subsec:fiqa}

FIQA approaches can be categorized into four groups: \textbf{(1) Label-generation approaches} train regression networks using quality labels from various sources. FaceQnet \cite{faceqnetv1} uses the comparison score between the sample and corresponding ICAO-compliant sample as the quality label, while SDD-FIQA \cite{SDDFIQA} employs distribution distances. RankIQ \cite{RANKIQ_FIQA} adopts a learning-to-rank strategy, training models to predict quality rankings based on FR performance metrics across different datasets. A limitation of these approaches is that they often decouple FIQA from FR, typically employing shallower networks that don't extract comprehensive facial features. \textbf{(2) Non-FR model approaches} include DifFIQA \cite{10449044}, which leverages diffusion models to assess embedding robustness under different conditions, and eDifFIQA \cite{babnikTBIOM2024}, which distills this approach into a lighter model for faster inference. While these methods can achieve high accuracy, they incur significant computational costs. \textbf{(3) Pre-trained FR analysis approaches} operate on fixed FR models without requiring additional training. SER-FIQ \cite{SERFIQ} measures embedding stability under dropout perturbations by evaluating embedding consistency with varied dropout patterns. GraFIQs \cite{grafiqs} uses gradient magnitudes during backpropagation to evaluate sample alignment with the FR model's objective. FaceQAN \cite{FaceQAN} estimates quality by quantifying adversarial robustness. \textbf{(4) FR-integrated approaches} directly incorporate quality assessment into the FR training process. MagFace \cite{MagFace} links quality scores to embedding magnitudes through regularized training. PFE \cite{PFE_FIQA} models embeddings as Gaussian distributions with uncertainty representing quality. CR-FIQA \cite{boutros_2023_crfiqa} estimates quality by predicting a sample's relative classifiability within the embedding space. ViT-FIQA \cite{atzori2025vitfiqaassessingfaceimage} extended standard ViT backbones with a learnable quality token designed to predict utility scores for face images. These FR-integrated approaches have consistently achieved top rankings in SOTA evaluations \cite{MagFace,boutros_2023_crfiqa,atzori2025vitfiqaassessingfaceimage}.

Among these categories, training-free methods, i.e. Pre-trained FR analysis approaches, offer the advantage of immediate applicability to pre-trained models without modification or fine-tuning. Our \ourmethod belongs to this category, requiring no additional training beyond the standard FR model. As summarized in Table \ref{tab:difference_methods}, existing training-free approaches rely on either multiple forward passes \cite{SERFIQ} or backpropagation \cite{grafiqs, FaceQAN}. In contrast, we exploit the hierarchical nature of ViT processing by analyzing the stability of patch representations across intermediate transformer blocks, requiring only a single forward pass without backpropagation. This design makes \ourmethod the only method using just a single forward pass among training-free FIQA methods while providing a novel perspective on quality assessment through cross-block feature stability analysis.

\section{Methodology}
\label{sec:methodology}

As discussed in Section \ref{subsec:vit_blocks}, research on ViT has demonstrated that transformer blocks refine features iteratively with high inter-block similarity \cite{raghu2021vit}, where each block produces slight refinements in the representations rather than complete transformations. This smooth feature evolution trajectory motivates our approach: we hypothesize, which we validate later in this section, that high-quality face images maintain stable patch representations across transformer blocks, while low-quality images exhibit larger changes due to quality-degrading factors such as blur, occlusion, or poor illumination.  We begin by formalizing the ViT architecture to establish the mathematical foundation for our quality assessment framework.

\textbf{Preliminaries on ViT:} Consider a ViT architecture \cite{DBLP:conf/iclr/DosovitskiyB0WZ21}, as shown in Figure \ref{fig:pipeline}. Given an input face image $\mathbf{I} \in \mathbb{R}^{H \times W \times 3}$ (height $H$, width $W$, RGB channels), the image is divided into non-overlapping patches of size $P \times P$, resulting in $N = \frac{HW}{P^2}$ patches. Each patch is linearly projected to an embedding of dimension $D$:\vspace{-5mm}

\begin{equation}
    \mathbf{z}_0 = [\mathbf{Y}\mathbf{p}_1 + \mathbf{b}; \mathbf{Y}\mathbf{p}_2 + \mathbf{b}; \ldots; \mathbf{Y}\mathbf{p}_N + \mathbf{b}] + \mathbf{E}_{pos},
    \vspace{-1mm}
\end{equation}
where $\mathbf{Y} \in \mathbb{R}^{D \times (P^2 \cdot 3)}$ is the patch embedding projection matrix, $\mathbf{p}_i \in \mathbb{R}^{P^2 \cdot 3}$ is the $i$-th flattened patch, $\mathbf{b} \in \mathbb{R}^{D}$ is the bias term,  and $\mathbf{E}_{pos} \in \mathbb{R}^{N \times D}$ are learnable positional embeddings. The embedded patches are processed through $L$ transformer blocks. Each block $\ell \in \{0, \ldots, L-1\}$ applies multi-head self-attention (MSA) followed by a multi layer perceptron (MLP) with residual connections:\vspace{-3mm}

\begin{equation}
\begin{aligned}
    \mathbf{z}'_\ell &= \text{MSA}(\text{LN}(\mathbf{z}_{\ell-1})) + \mathbf{z}_{\ell-1}, \\
    \mathbf{z}_\ell &= \text{MLP}(\text{LN}(\mathbf{z}'_\ell)) + \mathbf{z}'_\ell,
    \label{eq:msa_mlp}
\end{aligned}\vspace{-1mm}
\end{equation}
where LN denotes Layer Normalization and $\mathbf{z}_\ell \in \mathbb{R}^{N \times D}$ contains refined patch representations at block $\ell$. The residual connections (addition operations in Equation \ref{eq:msa_mlp}) maintain high similarity between blocks through influential skip connections that propagate feature information forward \cite{raghu2021vit}. The MSA mechanism at block $\ell$ computes query $\mathbf{Q}_\ell$, key $\mathbf{K}_\ell$, and value $\mathbf{V}_\ell$ matrices from the input, then applies scaled dot-product attention across $H$ heads:\vspace{-5mm}

\begin{equation}
    \text{MSA}(\mathbf{z}_{\ell-1}) = \text{Concat}(\text{head}_{\ell,1}, \ldots, \text{head}_{\ell,H})\mathbf{W}^O_\ell,
    \vspace{-1mm}
\end{equation}
where each attention head $h \in \{1, \ldots, H\}$ at block $\ell$ computes:\vspace{-6mm}

\begin{equation}
    \text{head}_{\ell,h} = \text{softmax}\left(\frac{\mathbf{Q}_{\ell,h}\mathbf{K}_{\ell,h}^\top}{\sqrt{D/H}}\right)\mathbf{V}_{\ell,h},
    \label{eq:attention}
    \vspace{-1mm}
\end{equation}
with $\mathbf{Q}_{\ell,h}, \mathbf{K}_{\ell,h}, \mathbf{V}_{\ell,h} \in \mathbb{R}^{N \times (D/H)}$ and $\mathbf{W}^O_\ell \in \mathbb{R}^{D \times D}$ as the output projection. The attention matrix $\mathbf{A}_{\ell,h} = \text{softmax}\left(\frac{\mathbf{Q}_{\ell,h}\mathbf{K}_{\ell,h}^\top}{\sqrt{D/H}}\right) \in \mathbb{R}^{N \times N}$ captures pairwise patch relationships at block $\ell$ and head $h$, where $\mathbf{A}_{\ell,h}^{(j,p)}$ represents the attention weight from patch $j$ to patch $p$.

\textbf{\ourmethod:} To capture the stability of this refinement process, we measure how patch embeddings evolve across intermediate transformer blocks. Let $\mathcal{T} = \{t_0, t_1, \ldots, t_{T-1}\} \subseteq \{0, 1, \ldots, L-1\}$ denote a selected subset of $T$ transformer blocks from which we extract intermediate representations, where $t_i + 1 = t_{i+1}$ always holds true. For each selected block $t_i \in \mathcal{T}$, we extract the patch embeddings $\mathbf{z}_{t_i} \in \mathbb{R}^{N \times D}$ and apply $L_2$ normalization to focus on directional changes rather than magnitude variations:\vspace{-5mm}

\begin{equation}
    \hat{\mathbf{z}}_{t_i}^{(p)} = \frac{\mathbf{z}_{t_i}^{(p)}}{\|\mathbf{z}_{t_i}^{(p)}\|_2},
    \vspace{-1mm}
\end{equation}
where $\mathbf{z}_{t_i}^{(p)} \in \mathbb{R}^D$ denotes the embedding vector of patch $p$ at block $t_i$, and $\hat{\mathbf{z}}_{t_i}^{(p)}$ is the unit-norm normalized embedding, which are illustrated as green blocks in Figure \ref{fig:pipeline}. This normalization ensures that we measure the angular change in feature representations, which is more robust to scale variations across different blocks.

For each patch $p \in \{1, \ldots, N\}$, we quantify the instability by computing the Euclidean distance between normalized embeddings from consecutive selected blocks:\vspace{-3mm}

\begin{equation}
    d_{t_i, t_{i+1}}^{(p)} = \|\hat{\mathbf{z}}_{t_i}^{(p)} - \hat{\mathbf{z}}_{t_{i+1}}^{(p)}\|_2,
    \vspace{-1mm}
\end{equation}
for $i \in \{0, \ldots, T-2\}$, where $d_{t_i, t_{i+1}}^{(p)}$ is the inter-block distance for patch $p$, shown as purple blocks in Figure \ref{fig:pipeline}. To obtain a comprehensive measure of patch stability across the entire refinement trajectory, we average these distances:\vspace{-3mm}

\begin{equation}
    \bar{d}^{(p)} = \frac{1}{T-1} \sum_{i=0}^{T-2} d_{t_i, t_{i+1}}^{(p)},
    \label{eq:avg_distance}
    \vspace{-1mm}
\end{equation}
where $\bar{d}^{(p)}$ is the average cross-block distance for patch $p$. This directly reflects how much a patch embedding changes as it propagates through the transformer blocks. To convert these distance measurements into interpretable quality scores, we apply a transformation that maps the continuous distance values to a bounded quality range:\vspace{-3mm}

\begin{equation}
    q^{(p)} = \frac{2}{1 + \exp(\alpha \cdot \bar{d}^{(p)})},
    \label{eq:patch_quality}
    \vspace{-1mm}
\end{equation}
where $\alpha > 0$ is a scaling parameter and $q^{(p)} \in (0, 1]$ is the quality score for patch $p$. Patch qualities (Blue blocks in Figure \ref{fig:pipeline}) obtained from patch distances (Purple blocks in Figure \ref{fig:pipeline}) through Equation \ref{eq:avg_distance}, and Equation \ref{eq:patch_quality}. This formulation maps smaller distances (stable patch representations) to quality scores approaching 1, and larger distances (unstable representations) to scores approaching 0, providing a smooth, monotonic mapping.

Having established patch-level quality scores, we now address the challenge of obtaining a single image-level quality estimate $Q \in (0, 1]$. Since different facial regions may exhibit varying degrees of quality degradation or utility for the FR task, we explore two aggregation strategies. First, we consider uniform aggregation that treats all patches equally:\vspace{-6mm}

\begin{equation}
    Q_{\text{uniform}} = \frac{1}{N} \sum_{p=1}^{N} q^{(p)}.
    \label{eq:uniform_agg}
    \vspace{-1mm}
\end{equation}

While this approach is simple, it does not account for the fact that certain facial regions (e.g., eyes, nose) may be more critical for recognition than others (e.g., background patches). To incorporate this spatial importance, we leverage the self-attention mechanism inherent in ViTs. We compute attention-based weights from the last transformer block ($\ell = L-1$):\vspace{-3mm}

\begin{equation}
    w^{(p)} = \frac{\sum_{h=1}^{H} \sum_{j=1}^{N} \mathbf{A}_{L-1,h}^{(j,p)}}{\sum_{p'=1}^{N} \sum_{h=1}^{H} \sum_{j=1}^{N} \mathbf{A}_{L-1,h}^{(j,p')}},
    \label{eq:attention_weights}
    \vspace{-1mm}
\end{equation}
where $\mathbf{A}_{L-1,h} \in \mathbb{R}^{N \times N}$ is the attention matrix of head $h \in \{1, \ldots, H\}$ at the last block, $\mathbf{A}_{L-1,h}^{(j,p)}$ is the attention weight from patch $j$ to patch $p$, and $w^{(p)} \in [0,1]$ is the normalized importance weight for patch $p$ with $\sum_{p=1}^{N} w^{(p)} = 1$. These weights are illustrated as yellow boxes in Figure \ref{fig:pipeline}, and this weighting scheme captures how much each patch is attended to during the recognition process. The attention-weighted quality score is then:\vspace{-4mm}

\begin{equation}
    Q_{\text{weighted}} = \sum_{p=1}^{N} w^{(p)} \cdot q^{(p)}.
    \label{eq:weighted_agg}
        \vspace{-2mm}
\end{equation}

The complete \ourmethod operates in a single forward pass through a pre-trained ViT model: it extracts intermediate patch representations at selected transformer blocks, computes normalized cross-block distances for each patch according to Equation \ref{eq:avg_distance}, transforms these distances to patch quality scores via Equation \ref{eq:patch_quality}, and aggregates them to an image-level score using either Equation \ref{eq:uniform_agg} or Equation \ref{eq:weighted_agg}. Critically, \ourmethod requires no additional training, no backpropagation, and no architectural modifications, enabling immediate deployment on any pre-trained ViT model while maintaining computational efficiency.

\textbf{Empirical Validation of \ourmethod:} We analyzed all 550,000 images from SynFIQA \cite{mrfiqa}, a quality-controlled synthetic dataset produced through a two-stage pipeline based on stable diffusion with controllable 3D facial parameters, dual text prompts for occlusion, and post-processing for blur and downsampling. The dataset contains 5,000 identities, each with 10 reference images and 100 degraded variants (10 per reference), organized into 11 quality groups. As shown in Figure \ref{fig:synfiqa}, we computed mean L2 distances between corresponding patch embeddings from consecutive ViT-B blocks across these quality groups. From this analysis, shown in Figure \ref{fig:synfiqa}, we see that lower consecutive-block distances systematically correspond to higher ground-truth quality across most block transitions, where the higher-quality groups (right side of each boxplot, representing better image quality) exhibit progressively lower average L2 distances compared to those for lower-quality groups (left side), with the inset for Block 11 $\leftrightarrow$ 12 explicitly illustrating this quality gradient (low $\rightarrow$ high) and showing how distances decrease as quality improves, thereby providing a clear measure of quality discriminability. This empirical evidence demonstrates that patch embedding stability across transformer blocks serves as an indicator of face image quality.

\section{Experimental Setup}
\label{sec:setup}

We utilized four pre-trained ViT models for FR, which are ViT-S/WebFace4M/Adaface \cite{DBLP:conf/cvpr/Kim0L22}, ViT-B/WebFace4M/Adaface \cite{DBLP:conf/cvpr/Kim0L22}, ViT-B/WebFace12M/Adaface \cite{DBLP:conf/cvpr/Kim0L22}, FRoundation ViT-B/16 \cite{DBLP:journals/ivc/ChettaouiDB25}, and one pre-trained foundation model, CLIP ViT-B/16 \cite{radford2021learningtransferablevisualmodels} to showcase our methods applicability to any pre-trained ViT model. We conducted extensive experiments across eight benchmark datasets: LFW \cite{LFWTech}, AgeDB-30 \cite{agedb}, CFP-FP \cite{cfp-fp}, CALFW \cite{CALFW}, Adience \cite{Adience}, CPLFW \cite{CPLFWTech}, XQLFW \cite{XQLFW}, and IJB-C \cite{ijbc}. Performance was measured using Error-versus-Discard Characteristic (EDC) curves \cite{GT07}, which assess the impact of discarding low-quality face images on face verification performance and quantify how verification errors decrease as low-quality samples are progressively removed. The False Non-Match Rate (FNMR) was evaluated at fixed False Match Rate (FMR) thresholds \cite{iso_metric}, specifically at $1e-3$ (recommended for border control by Frontex \cite{frontex2015best}) and $1e-4$ (for higher security applications). Additionally, we reported the Area Under the Curve (AUC) and partial AUC (pAUC) of the EDC curves to quantify verification performance across rejection rates. The pAUC \cite{10449044, babnikTBIOM2024, DBLP:journals/tbbis/SchlettRTB24} measures performance up to a 25\% rejection rate. To thoroughly examine the impact of our FIQA approaches across different FR architectures, we evaluated performance using four SOTA CNN-based models: ArcFace \cite{deng2019arcface}, ElasticFace \cite{elasticface}, MagFace \cite{MagFace}, and CurricularFace \cite{curricularFace}. All evaluations were conducted under cross-model settings, where the models used to evaluate FIQA were different from those used to extract face feature representations.

\begin{table*}[!ht]
\centering
\setlength{\tabcolsep}{3pt}
\caption{Ablation studies analyzing four design choices: dataset generalization (WebFace4M, WebFace12M, CLIP, FRoundation), architecture depth (ViT-S vs ViT-B), block depth trade-offs (4-24 blocks), and aggregation strategies (uniform vs attention-weighted). Results show optimal performance at 12-20 blocks with last-block attention weighting. Mean pAUC-EDC computed across seven benchmarks at FMR=$1e-3$ and $1e-4$. Best per study in bold.}
\label{tab:ablation_pauc}
\vspace{-3mm}
\resizebox{\textwidth}{!}{
\begin{tabular}{|c|c|c|
cc|cc|cc|cc|cc|cc|cc|cc|}
\hline
\multirow{2}{*}{\textbf{Study}} & \multirow{2}{*}{\textbf{Method}}  & \multirow{2}{*}{\textbf{Blocks}}
& \multicolumn{2}{c|}{Adience \cite{Adience}} 
& \multicolumn{2}{c|}{AgeDB-30 \cite{agedb}} 
& \multicolumn{2}{c|}{CFP-FP \cite{cfp-fp}} 
& \multicolumn{2}{c|}{LFW \cite{LFWTech}} 
& \multicolumn{2}{c|}{CALFW \cite{CALFW}} 
& \multicolumn{2}{c|}{CPLFW \cite{CPLFWTech}} 
& \multicolumn{2}{c|}{XQLFW \cite{XQLFW}} 
& \multicolumn{2}{c|}{Mean pAUC-EDC} \\

\cline{4-19}
 & & 
 & $1e{-3}$ & $1e{-4}$ 
 & $1e{-3}$ & $1e{-4}$ 
 & $1e{-3}$ & $1e{-4}$ 
 & $1e{-3}$ & $1e{-4}$ 
 & $1e{-3}$ & $1e{-4}$ 
 & $1e{-3}$ & $1e{-4}$ 
 & $1e{-3}$ & $1e{-4}$ 
 & $1e{-3}$ & $1e{-4}$ \\

\hline
\multirow{4}{*}{Dataset}
 & ViT-B - WebFace4M     & 0-23 
 &0.0102 & \textbf{0.0230} & 0.0085 & \textbf{0.0126} & 0.0065 & \textbf{0.0095} & 0.0008 & 0.0009 & \textbf{0.0196} & \textbf{0.0215} & 0.0233 & \textbf{0.0357} & \textbf{0.1267} & \textbf{0.1426} & \textbf{0.0279} & \textbf{0.0351} \\
 & ViT-B - WebFace12M     & 0-23
 & \textbf{0.0100}	&0.0234	&\textbf{0.0079}	&0.0129&	\textbf{0.0063}&	0.0101	&\textbf{0.0006}	&\textbf{0.0008}	&0.0209&	0.0230&	\textbf{0.0228}	&0.0359	&0.1275	&0.1514	&0.0280&	0.0368 \\
 & CLIP  & 0-11
 & 0.0156&	0.0349&	0.0092	&0.0141&	0.0098&	0.0139	&0.0008&	0.0009&	0.0200	&0.0224	&0.0465	&0.0633	&0.1522&	0.1697&	0.0363	&0.0456\\
 & FRoundation  & 0-11 
 & 0.0154	&0.0356&	0.0099&	0.0150&	0.0093	&0.0130	&0.0007	&0.0009	&0.0205&	0.0230&	0.0410	&0.0640	&0.1525	&0.1699	&0.0356	&0.0459 \\
\hline
\end{tabular}}\vspace{-6mm}
\end{table*}

\begin{table*}[ht!]
    \begin{center}
    \caption{The pAUCs of EDC achieved by our method and the SOTA methods under different experimental settings. The notions of $1e-3$ and $1e-4$ indicate the value of the fixed FMR at which the EDC curves (FNMR vs.~reject) were calculated. The results are compared to three IQA and twelve FIQA approaches. The XQLFW dataset uses SER-FIQ (marked with *) as the FIQ labeling method.}
    \label{tab:sota_pauc}
    \vspace{-3mm}
    \resizebox{0.99\textwidth}{!}{
\begin{tabular}{|c|cl|cc|cc|cc|cc|cc|cc|cc|cc|}
    \hline
\multirow{3}{*}{FR} & \multicolumn{2}{c|}{\multirow{3}{*}{Method}} 
& \multicolumn{2}{c|}{Adience} & \multicolumn{2}{c|}{AgeDB-30} & \multicolumn{2}{c|}{CFP-FP} 
& \multicolumn{2}{c|}{LFW} & \multicolumn{2}{c|}{CALFW} & \multicolumn{2}{c|}{CPLFW} 
& \multicolumn{2}{c|}{XQLFW} & \multicolumn{2}{c|}{IJB-C} \\ 
 & & 
 & \multicolumn{2}{c|}{\cite{Adience}} & \multicolumn{2}{c|}{\cite{agedb}} & \multicolumn{2}{c|}{\cite{cfp-fp}} 
 & \multicolumn{2}{c|}{\cite{LFWTech}} & \multicolumn{2}{c|}{\cite{CALFW}} & \multicolumn{2}{c|}{\cite{CPLFWTech}} 
 & \multicolumn{2}{c|}{\cite{XQLFW}}  & \multicolumn{2}{c|}{\cite{ijbc}} \\ \cline{4-19}
                                                                                    &                                                   &  & $1e{-3}$ & $1e{-4}$ & $1e{-3}$ & $1e{-4}$  & $1e{-3}$ & $1e{-4}$  & $1e{-3}$ & $1e{-4}$  & $1e{-3}$ & $1e{-4}$ & $1e{-3}$ & $1e{-4}$ & $1e{-3}$ & $1e{-4}$ & $1e{-3}$ & $1e{-4}$ \\ \hline
     \hline 
     \multirow{16}{*}{\rotatebox[origin=c]{90}{ArcFace\cite{deng2019arcface}}}      & \multirow{3}{*}{\rotatebox[origin=c]{90}{IQA}}    & \multicolumn{1}{|l|}{BRISQUE\cite{BRISQE_IQA}} & 0.0143  & 0.0333  & 0.0096  & 0.0146  & 0.0095  & 0.0136  & 0.0009  & 0.0010  & 0.0200  & 0.0225  & 0.0501  & 0.0638  & 0.1512  & 0.1689  & 0.0072	&	0.0113     \\ 
                                                                                    &                                                   & \multicolumn{1}{|l|}{RankIQA\cite{liu2017rankiqa}} & 0.0124  & 0.0302  & 0.0087  & 0.0141  & 0.0088  & 0.0135  & 0.0009  & 0.0010  & 0.0209  & 0.0234  & 0.0506  & 0.0658  & 0.1532  & 0.1709  & 0.0071	&	0.0112\\
\hline 
    \end{tabular}}
    \end{center}
    \vspace{-6mm}
    \end{table*}

\begin{figure}[t]
    \centering
    \setlength{\tabcolsep}{1pt}
    \renewcommand{\arraystretch}{1.1}
    \resizebox{0.45\textwidth}{!}{
        \includegraphics[width=0.45\textwidth]{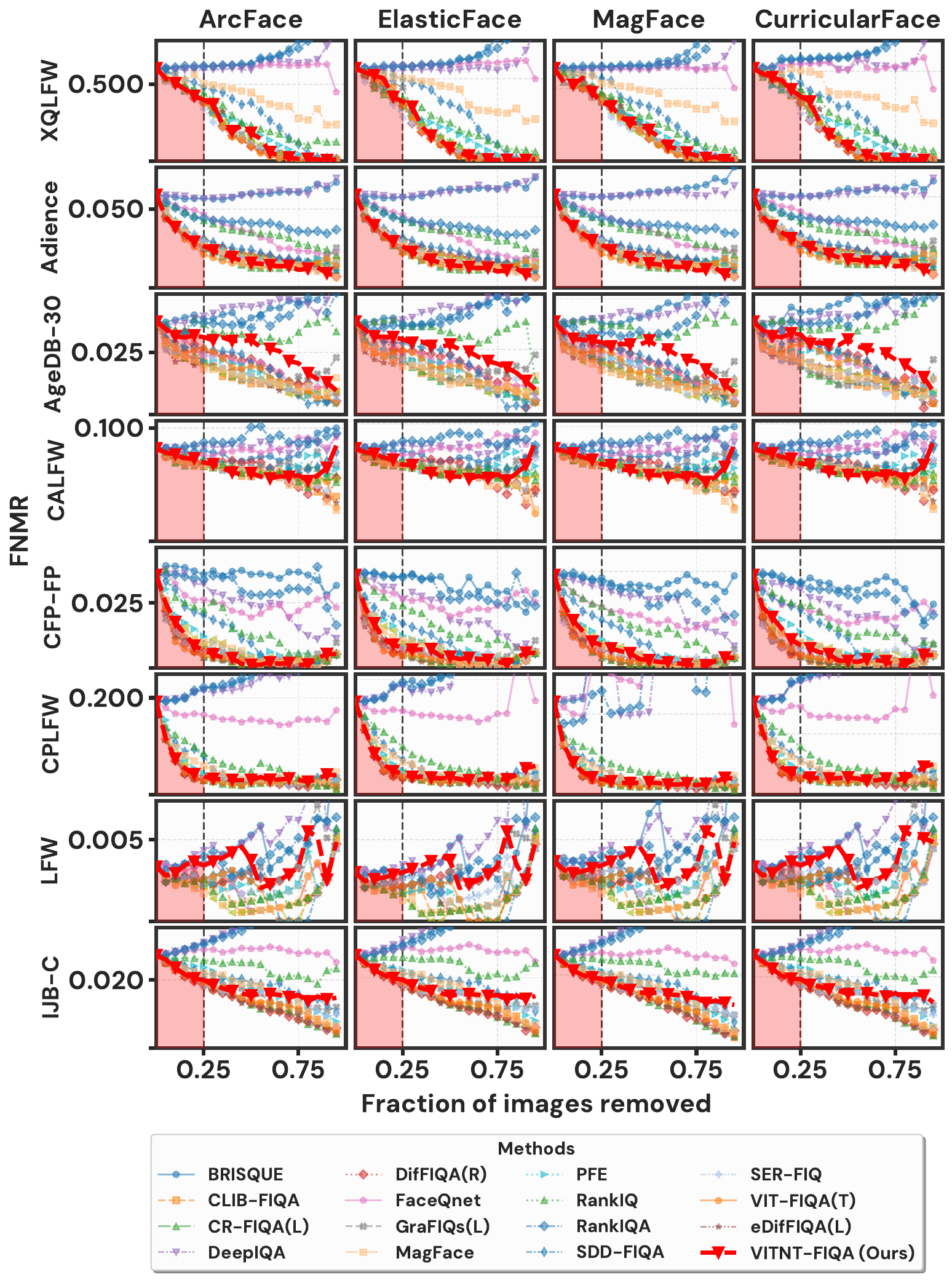}}
    \caption{Error-versus-Discard Characteristic (EDC) curves for FNMR@FMR=$1e-3$ of our proposed method in comparison to SOTA. Results shown on eight benchmark datasets: LFW \cite{LFWTech}, AgeDB-30 \cite{agedb}, CFP-FP \cite{cfp-fp}, CALFW \cite{CALFW}, Adience \cite{Adience}, CPLFW \cite{CPLFWTech}, XQLFW \cite{XQLFW}, and IJB-C \cite{ijbc}, using ArcFace \cite{deng2019arcface}, ElasticFace \cite{elasticface}, MagFace \cite{MagFace}, and CurricularFace \cite{curricularFace} FR models. Our method \ourmethod is marked with the red line.}
    \label{fig:fnmr3}\vspace{-6mm}
\end{figure}

\vspace{-3mm}
\section{Results}
\label{sec:results}

We conduct comprehensive ablation studies (Table \ref{tab:ablation_pauc}) to analyze the impact of various design choices, followed by comparison with SOTA methods (Table \ref{tab:sota_pauc} and Figure \ref{fig:fnmr3}).

\subsection{Ablation Studies}

\textbf{Dataset Study.} We evaluate \ourmethod across different pre-trained models with comparable ViT-B and ViT-B/16 architectures trained on varying datasets. The ViT-B/WebFace4M/AdaFace \cite{DBLP:conf/cvpr/Kim0L22} and ViT-B/WebFace12M/AdaFace \cite{DBLP:conf/cvpr/Kim0L22} models, both trained specifically for FR tasks, achieve nearly identical performance (mean pAUC-EDC of 0.0279/0.0351 vs 0.0280/0.0368 at FMR=$1e-3$/$1e-4$), demonstrating that \ourmethod generalizes well on small and large datasets. Notably, CLIP ViT-B/16 \cite{radford2021learningtransferablevisualmodels}, a foundation model not trained for FR at all, yields worse results (0.0363/0.0456), showing that cross-block patch embedding stability correlates, to some degree, with face quality even in models without FR-specific training. FRoundation ViT-B/16 \cite{DBLP:journals/ivc/ChettaouiDB25}, which adapts CLIP for FR through LoRA layers while retaining multi-task capabilities, achieves similar performance (0.0356/0.0459) to CLIP. This demonstrates that \ourmethod is immediately applicable to any pre-trained ViT model without requiring FR-specific fine-tuning, making it highly versatile for deployment across different model families and training paradigms. However, we observe that \ourmethod performs better with FR-specific-trained models, as FIQA is highly coupled with the FR task \cite{boutros_2023_crfiqa}.

\textbf{Architecture Study:} Comparing ViT-S (12 blocks) and ViT-B (24 blocks) trained on the same WebFace4M dataset reveals minimal performance differences (0.0273/0.0359 vs 0.0279/0.0351). While ViT-S shows marginal advantages on certain datasets (e.g., CFP-FP, CALFW), ViT-B performs better on others (e.g., Adience, XQLFW). This indicates that \ourmethod effectively captures quality-relevant information regardless of network depth, as both architectures exhibit the smooth feature refinement trajectory that our method exploits.

\textbf{Block Depth Study:} We systematically vary the number of blocks used (4, 8, 12, 16, 20, 24) for ViT-B/WebFace4M/AdaFace \cite{DBLP:conf/cvpr/Kim0L22} to understand the trade-off between computational efficiency and performance. Using only 4 blocks (0-3) yields the highest computational savings but weaker performance (0.0297/0.0379), particularly on CPLFW. Performance steadily improves as more blocks are included, with optimal results achieved at 16 blocks (0.0262/0.0336) and 20 blocks (0.0266/0.0338). Interestingly, using all 24 blocks (0.0279/0.0351) slightly degrades performance compared to 16-20 blocks. This finding indicates that practitioners can achieve near-optimal performance using only blocks 0-15, reducing computational costs. The sweet spot at 12-20 blocks balances efficiency and effectiveness when the uniform aggregation strategy is used, making it practical for resource-constrained deployments while maintaining competitive quality assessment.

\textbf{Attention-Weighting Study:} We compare uniform patch aggregation (Equation \ref{eq:uniform_agg}) with attention-weighted aggregation (Equation \ref{eq:weighted_agg}) using weights from either the last block or averaged across all blocks. The uniform baseline (using blocks 0-23) achieves 0.0279/0.0351. Attention-weighting from the last block consistently improves performance across different depth configurations, with the best result at 12 blocks (0.0260/0.0334), demonstrating that not all patches contribute equally to quality assessment. Averaging attention weights across all blocks (0.0273/0.0342) performs comparably to last-block weighting, suggesting that the spatial importance patterns remain relatively stable throughout the network. This validates our design choice to leverage self-attention for patch importance weighting, the attention mechanism naturally identifies quality-critical regions. The consistent improvement from attention-weighting indicates that ViT's learned attention patterns align well with quality-relevant regions, providing a principled way to aggregate patch-level quality scores without additional supervision.

\vspace{-2mm}
\subsection{Comparison with State-of-the-Art}

Table \ref{tab:sota_pauc} presents comprehensive comparisons with SOTA across four FR models (ArcFace \cite{deng2019arcface}, ElasticFace \cite{elasticface}, MagFace \cite{MagFace}, CurricularFace \cite{curricularFace}) and eight benchmark datasets. Our \ourmethod demonstrates competitive performance with SOTA FIQA methods while offering distinct advantages in applicability.

\textbf{Training-Free Methods:} Compared to training-free methods (Table \ref{tab:difference_methods}), \ourmethod achieves performance on par with or better than SER-FIQ \cite{SERFIQ} and GraFIQs \cite{grafiqs} across multiple datasets and FR models. Notably, SER-FIQ requires 100 forward passes with stochastic dropout to measure embedding stability, while GraFIQs requires backpropagation to compute gradient magnitudes. In contrast, \ourmethod achieves comparable or superior results using only a single forward pass without backpropagation. On the challenging Adience dataset, \ourmethod achieves 0.0095/0.0226 (ArcFace), 0.0107/0.0209 (ElasticFace), 0.0097/0.0225 (MagFace), and 0.0084/0.0191 (CurricularFace) at FMR=$1e-3$/$1e-4$, consistently outperforming SER-FIQ (0.0102/0.0244, 0.0114/0.0227, 0.0107/0.0241, 0.0091/0.0207) and matching GraFIQs(L) (0.0093/0.0215, 0.0101/0.0203, 0.0097/0.0217, 0.0085/0.0186). On IJB-C, \ourmethod demonstrates robust performance (0.0058/0.0087 for ArcFace, 0.0055/0.0085 for ElasticFace) compared to SER-FIQ (0.0056/0.0087, 0.0054/0.0083) and GraFIQs(L) (0.0059/0.0089, 0.0056/0.0086).

\textbf{FR-Integrated Methods:} Among FR-integrated methods that require additional training, \ourmethod remains competitive with top-performing approaches across diverse evaluation scenarios. On Adience, our method consistently matches or approaches the performance of CR-FIQA(L) \cite{boutros_2023_crfiqa} (0.0095 vs 0.0097 for ArcFace, 0.0084 vs 0.0089 for CurricularFace at FMR=$1e-3$) and ViT-FIQA(T) \cite{atzori2025vitfiqaassessingfaceimage} (0.0095 vs 0.0089 for ArcFace, 0.0084 vs 0.0079 for CurricularFace), despite not requiring any custom loss functions or quality-specific training. On CPLFW, \ourmethod achieves 0.0200/0.0324 (ArcFace) and 0.0169/0.0292 (CurricularFace), performing comparably to CR-FIQA(L) (0.0190/0.0307, 0.0161/0.0283) and ViT-FIQA(T) (0.0191/0.0309, 0.0160/0.0281). Compared to diffusion-based methods DifFIQA \cite{10449044} and eDifFIQA \cite{babnikTBIOM2024}, which leverage generative models and incur high computational costs, \ourmethod provides similar or better performance across multiple datasets. For instance, on AgeDB-30 with MagFace, \ourmethod achieves 0.0084/0.0181 versus DifFIQA's 0.0084/0.0170 and eDifFIQA's 0.0065/0.0111, while on XQLFW with ElasticFace, \ourmethod achieves 0.1203/0.1450 compared to DifFIQA's 0.1138/0.1303 and eDifFIQA's 0.1195/0.1394, demonstrating competitive performance without requiring generative model training.

\textbf{EDC:} Figure \ref{fig:fnmr3} visualizes EDC curves at FMR=$1e-3$ across all datasets and FR models. Our method (red line) consistently tracks SOTA approaches across rejection rates, particularly on a challenging dataset like Adience. The curves demonstrate that \ourmethod effectively identifies low-quality samples, as more low-quality images are discarded, verification errors decrease steadily.

\vspace{-4mm}
\section{Conclusion}
\label{sec:conclusion}

We introduced \ourmethod, a training-free FIQA method that measures the stability of patch embedding evolution across intermediate ViT blocks to assess its utility of face image for FR. Our approach is grounded in the hypothesis that high-quality face images exhibit stable feature refinement trajectories across transformer blocks, while low quality images show erratic changes. By measuring L2 distances between normalized patch embeddings from consecutive blocks and aggregating them using attention-weighted schemes, \ourmethod produces quality scores in a single forward pass without requiring backpropagation, architectural modifications, or quality-specific training. Comprehensive evaluations across eight benchmarks and four FR models demonstrate that \ourmethod achieves competitive performance with state-of-the-art methods while offering distinct practical advantages. Our ablation studies reveal that: (1) the method generalizes across pre-trained models regardless of training data or even task specialization, (2) architecture depth has minimal impact on performance, (3) using a subset of encoder blocks provides optimal efficiency-performance trade-offs with computational savings when the uniform aggregation is used, and (4) attention-based patch weighting consistently improves quality assessment. The key contribution of \ourmethod lies in demonstrating that intermediate ViT representations contain quality-relevant information beyond serving as stepping stones to final embeddings. By exploiting the smooth feature refinement trajectory inherent to transformer architectures, our method provides a principled, efficient, and immediately applicable solution for face image quality assessment in modern recognition systems.

\vspace{-2mm}
\section*{Acknowledgment}
This research work has been funded by the German Federal Ministry of Education and Research and the Hessen State Ministry for Higher Education, Research and the Arts within their joint support of the National Research Center for Applied Cybersecurity ATHENE.

\clearpage
{
    \small
    \bibliographystyle{ieeenat_fullname}
    \bibliography{main}
}

\clearpage

\appendix
\section*{Supplementary Material}
This supplementary material provides comprehensive experimental results and detailed analysis of the \ourmethod method for face image quality assessment. The supplementary material is structured to address five fundamental research questions: (1) How do cross-block patch embedding distances correlate with image quality across different network depths? (2) Which block configurations provide optimal performance-efficiency trade-offs? (3) How does our method compare against existing state-of-the-art approaches across multiple evaluation metrics? (4) What visual patterns emerge in the ablation study EDC curves? (5) How does quality score distribution vary across different FIQA methods?

\section*{Tables: Quantitative Evidence}

\begin{itemize}
    \item \textbf{Table \ref{tab:block_ablation}}: Block window analysis comparing consecutive 6-block segments across ViT-B architecture. This systematic evaluation identifies which transformer block windows (early: 0-5, middle: 6-17, late: 18-23) capture the most quality-discriminative information. We include this analysis to demonstrate that \textit{early transformer blocks (0-5) achieve superior performance} across both AUC-EDC and pAUC-EDC metrics, providing empirical evidence that quality-relevant features emerge in initial processing stages rather than requiring full network depth.

    \item \textbf{Table \ref{tab:ablation_auc}}: Comprehensive ablation study presenting AUC-EDC performance for all four design choices (Dataset, Architecture, Block Depth, Attention-Weighting) across eight benchmark datasets using ArcFace as the face recognition model. Lower AUC-EDC values indicate better quality assessment performance. This systematic evaluation complements the pAUC-EDC analysis of the main paper, providing a complete picture of method performance across the full rejection rate spectrum (0-100\%) rather than just the partial area up to 25\% rejection. The AUC-EDC metric is essential to validate that our findings hold beyond the 25\% rejection threshold.

    \item \textbf{Table \ref{tab:sota_auc}}: State-of-the-art comparison presenting AUC-EDC values for our method against 15 competing approaches (3 IQA, 12 FIQA) across four face recognition models (ArcFace, ElasticFace, MagFace, CurricularFace). We provide this extensive comparison table in addition to the pAUC-EDC comparison, provided in the main paper, to establish the competitive advantage of our training-free approach across different evaluation metrics and operating points.
\end{itemize}

\section*{Figures: Visual Evidence and Insights}

\begin{figure}[ht]
    \centering
    \includegraphics[width=0.5\textwidth]{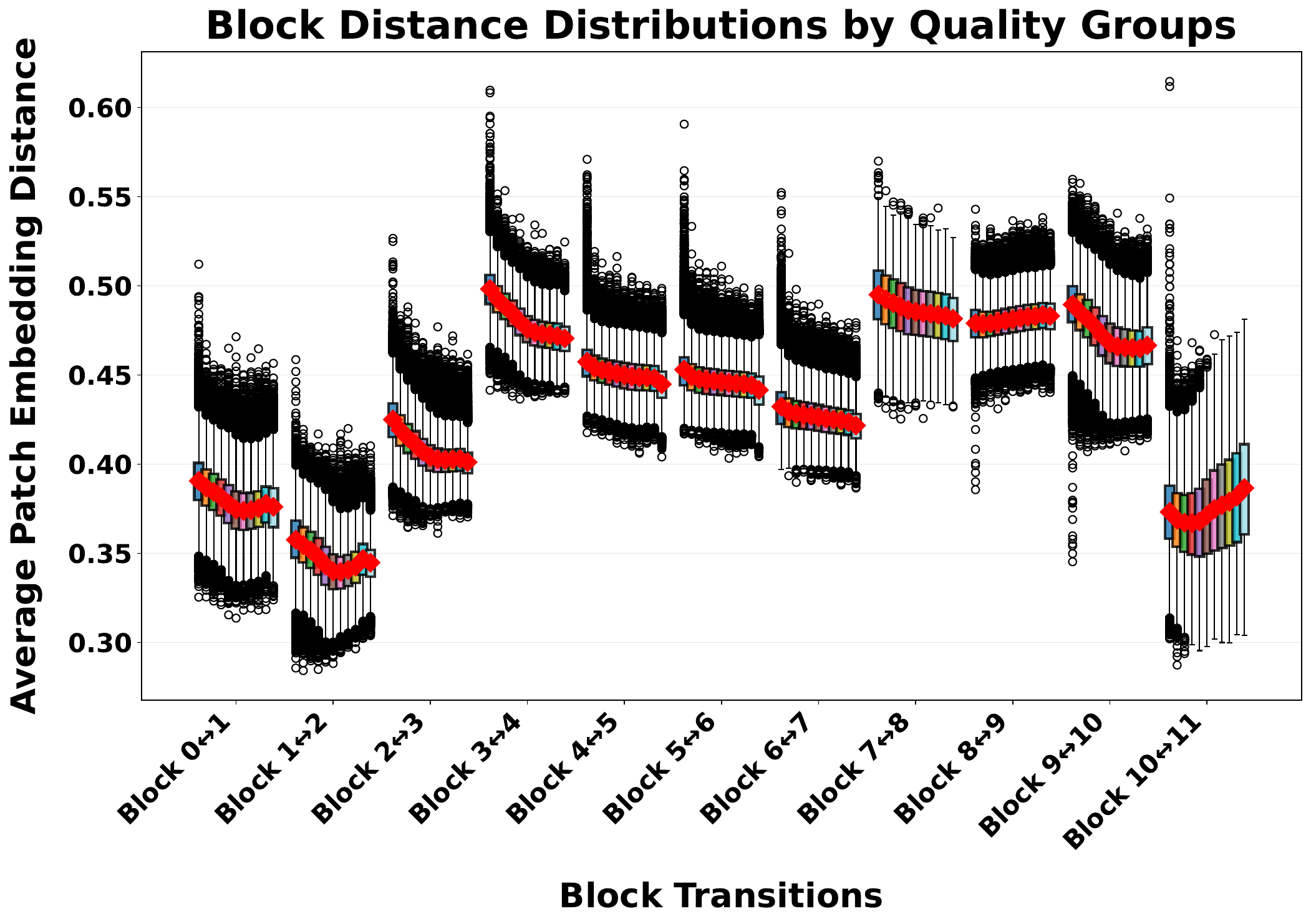}
    \caption{Boxplots of mean L2 distances between corresponding patch embeddings from consecutive ViT-S blocks computed for 11 quality groups, each having 0.5M images, from 5.5M images of SynFIQA \cite{mrfiqa}. Each box summarizes the distribution of average patch-embedding distances across images in a quality group, lower distances empirically correspond to higher ground-truth quality for most block transitions, i.e. the higher the quality, the lower the distance.}
    \label{fig:synfiqa_vits}
\end{figure}

\begin{figure*}[t]
    \centering
    \includegraphics[width=\textwidth]{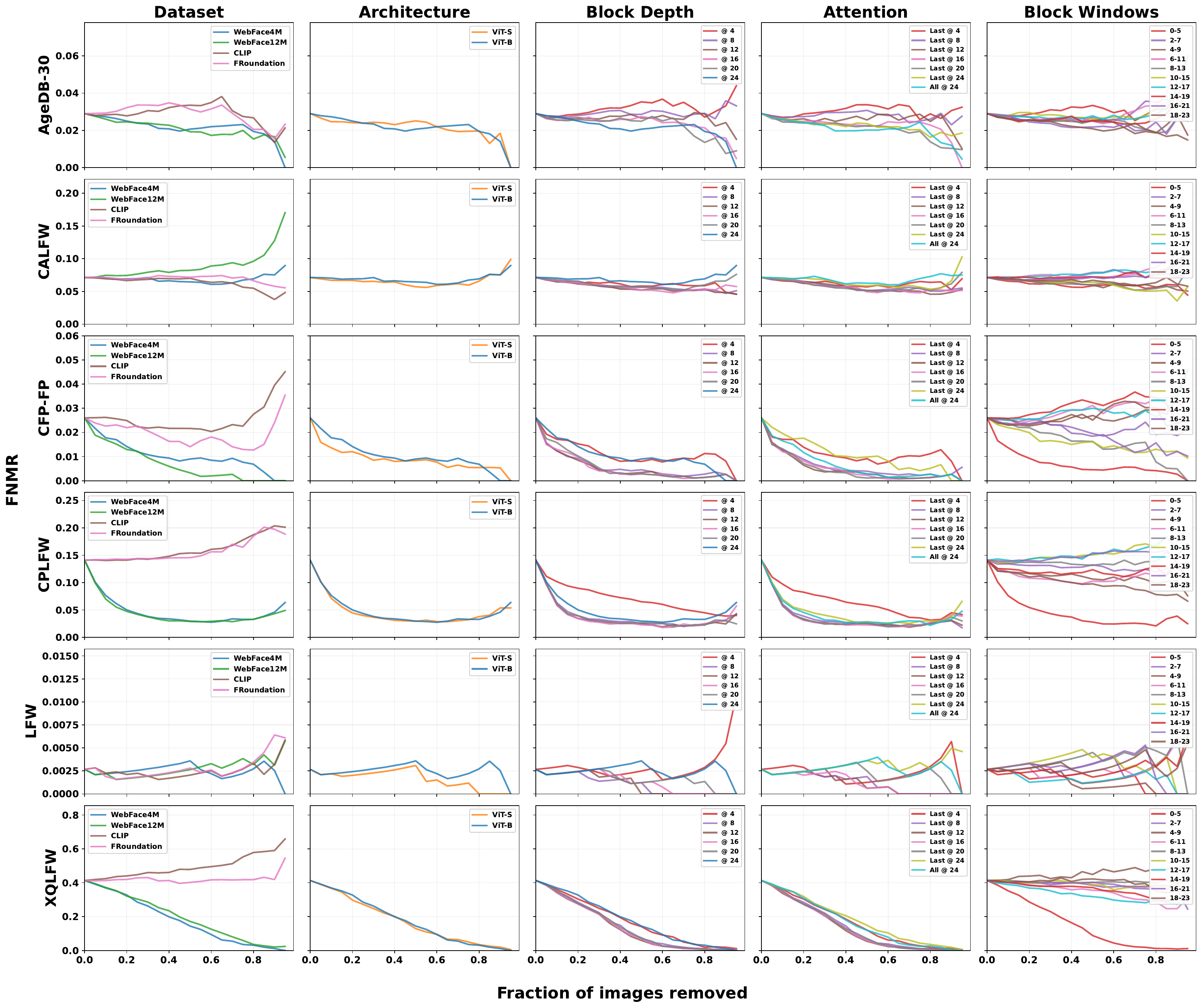}
    \caption{Comprehensive ablation analysis via Error-versus-Discard Characteristic (EDC) curves at FMR=$1e-2$. Each column represents one of five ablation studies: \textbf{Dataset} (generalization across WebFace4M, WebFace12M, CLIP, FRoundation), \textbf{Architecture} (ViT-S vs ViT-B depth comparison), \textbf{Block Depth} (computational trade-offs from 4 to 24 blocks), \textbf{Attention} (last-block vs all-blocks aggregation at varying depths), and \textbf{Block Windows} (consecutive 6-block segments from early to late network stages). Each row shows results on a different benchmark dataset (AgeDB-30, CALFW, CFP-FP, CPLFW, LFW, XQLFW). The Dataset study confirms cross-model generalization with FR-trained models (WebFace4M, WebFace12M) outperforming foundation models (CLIP, FRoundation). The Architecture study reveals minimal performance gap between ViT-S and ViT-B, validating depth-independence. The Block Depth study demonstrates that 12-20 blocks provide optimal efficiency-performance balance, with diminishing returns beyond 16 blocks. The Attention study shows consistent improvements from attention-weighting, particularly at 12-20 block depths. The Block Windows study reveals that early transformer blocks (0-5) capture the strongest quality signals. All curves use ArcFace for cross-model evaluation. Across all studies, FNMR decreases steadily as low-quality samples are discarded, validating \ourmethod's effectiveness in identifying quality-degraded images. The consistent color coding highlights method performance: WebFace4M-based configurations (blue) serve as the primary baseline across multiple studies.}
    \label{fig:erc_ablations_fnmr2}
\end{figure*}

\begin{figure*}[t]
    \centering
    \includegraphics[width=\textwidth]{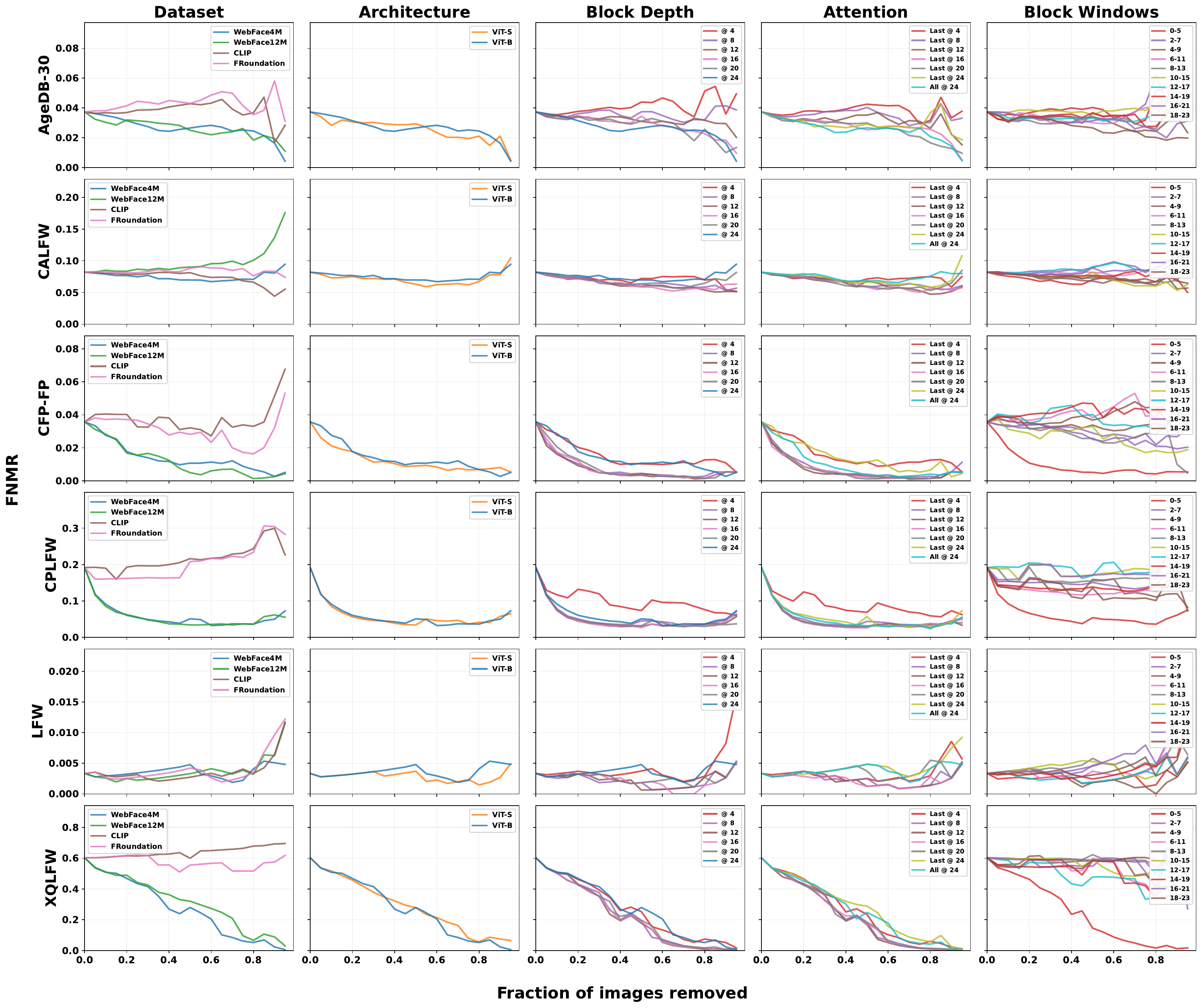}
    \caption{Comprehensive ablation analysis via Error-versus-Discard Characteristic (EDC) curves at FMR=$1e-3$ (Frontex-recommended threshold for border control applications). Layout identical to Figure~\ref{fig:erc_ablations_fnmr2}, similar conclusions are also drawn.}
    \label{fig:erc_ablations_fnmr3}
\end{figure*}

\begin{figure*}[t]
    \centering
    \includegraphics[width=\textwidth]{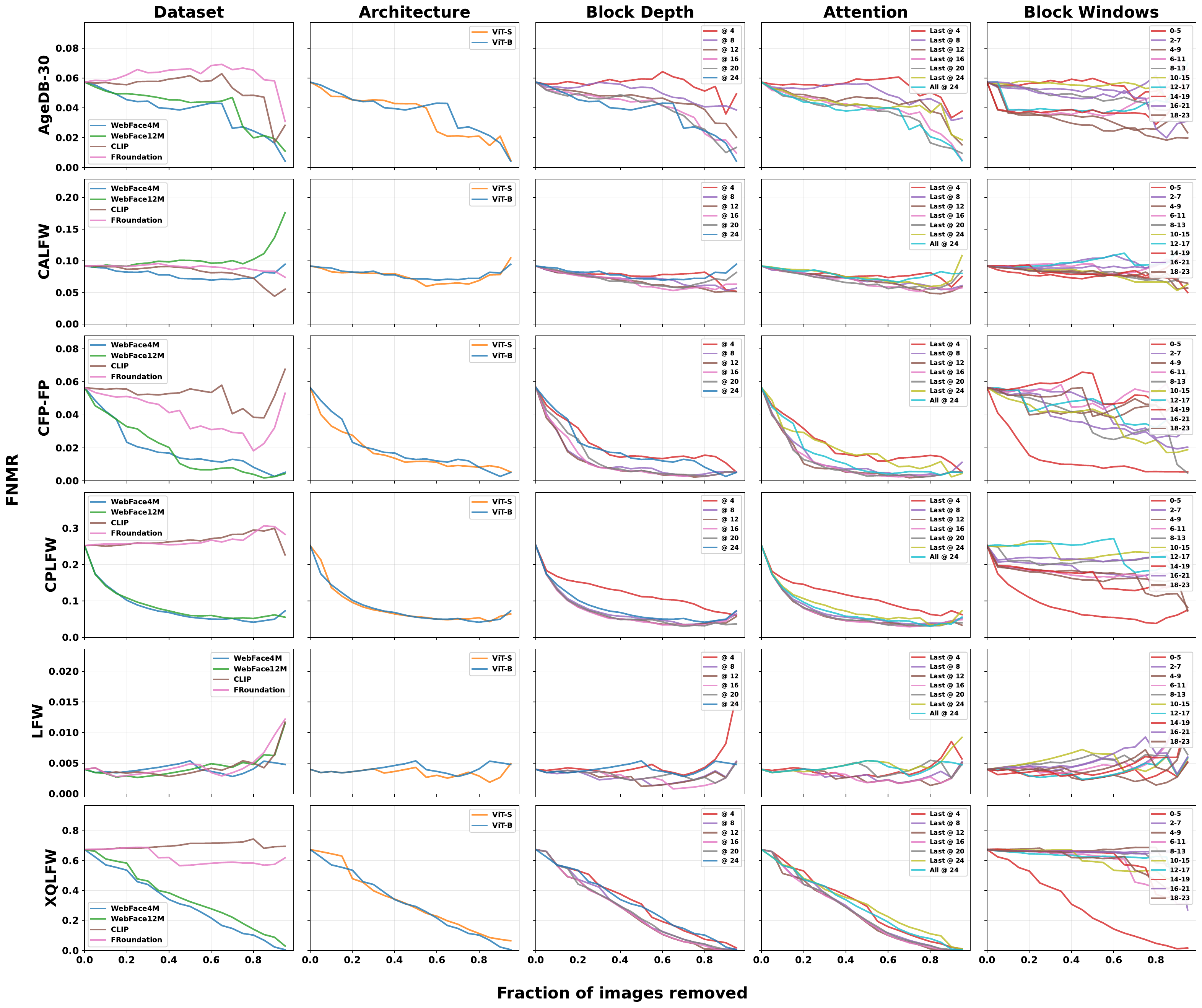}
    \caption{Comprehensive ablation analysis via Error-versus-Discard Characteristic (EDC) curves at FMR=$1e-4$ (high-security operating point). Layout identical to Figures~\ref{fig:erc_ablations_fnmr2} and~\ref{fig:erc_ablations_fnmr3}, similar conclusions are also drawn.}
    \label{fig:erc_ablations_fnmr4}
\end{figure*}

\begin{itemize}
    \item \textbf{Figure \ref{fig:synfiqa_vits}}: Boxplots of mean L2 distances between consecutive ViT-S patch embeddings across 11 quality groups from 5.5M SynFIQA images. This visualization complements the SynFIQA figure in the main paper (which shows ViT-B results) by demonstrating that \textit{our core hypothesis generalizes across different architecture scales}. The systematic decrease in cross-block distances with increasing ground-truth quality is evident in ViT-S (12 blocks) just as in ViT-B (24 blocks), confirming that patch embedding stability serves as a quality indicator regardless of network depth.

    \item \textbf{Figures \ref{fig:erc_ablations_fnmr2}, \ref{fig:erc_ablations_fnmr3}, \ref{fig:erc_ablations_fnmr4}}: Comprehensive ablation analysis via Error-versus-Discard Characteristic (EDC) curves at three security operating points (FMR=$1e-2$, $1e-3$, $1e-4$). \textit{The visual trends provide intuitive insights that complement the quantitative tables}: In the Dataset study column, blue/green curves (WebFace4M/WebFace12M) consistently lie below brown/pink curves (CLIP/FRoundation), visually confirming FR-specific training superiority. In the Architecture study, blue (ViT-B) and orange (ViT-S) curves run nearly parallel with minimal separation, demonstrating depth-independence. In the Block Depth study, we observe progressive downward shifts from red (4 blocks) to blue (24 blocks) up to 16 blocks, after which curves plateau or slightly rise, visually identifying the optimal 12-20 block sweet spot. In the Attention-Weighting study, we see a similar trend to Block Depth study, but with slightly lower curves. In the Block Windows study, the clear downward progression of the EDC curve in the early blocks (blocks 0-5), not seeing a similar downward progression for the others, visually confirms that quality discrimination concentrates in initial processing stages.

    \item \textbf{Figures \ref{fig:fnmr2}, \ref{fig:fnmr4}}: State-of-the-art comparison EDC curves at two additional FNMR@FMR thresholds ($1e-2$, $1e-4$) beyond the main paper's $1e-3$ threshold. These comprehensive comparisons across eight benchmark datasets and four face recognition models demonstrate that \textit{our method's competitiveness holds across multiple security requirements}.

    \item \textbf{Figure \ref{fig:quality_dist}}: Distribution of quality scores across evaluation benchmarks, comparing \ourmethod with SOTA methods. The normalized score distributions (range [0, 1]) reveal whether methods produce well-calibrated distributions that effectively rank sample quality or if they suffer from range compression that limits discriminative power.
\end{itemize}

\begin{table*}[ht!]
\centering
\setlength{\tabcolsep}{3pt}
\caption{Block-window analysis comparing quality-assessment performance across consecutive 6-block segments of ViT-B/WebFace4M. Early window (blocks 0–5) yields the strongest AUC-EDC and pAUC-EDC performance, indicating that initial feature refinements carry the most quality-discriminative signal. Mean metrics reported across seven benchmarks at FMR=$1\mathrm{e}{-3}$ and $1\mathrm{e}{-4}$. Best results per metric are in bold.}
\label{tab:block_ablation}
\vspace{-3mm}
\resizebox{\textwidth}{!}{\begin{tabular}{|c|c|c|
cc|cc|cc|cc|cc|cc|cc|cc|}
\hline
\multirow{2}{*}{\textbf{Metric}} & \multirow{2}{*}{\textbf{Method}}  & \multirow{2}{*}{\textbf{Blocks}}
& \multicolumn{2}{c|}{Adience \cite{Adience}} 
& \multicolumn{2}{c|}{AgeDB-30 \cite{agedb}} 
& \multicolumn{2}{c|}{CFP-FP \cite{cfp-fp}} 
& \multicolumn{2}{c|}{LFW \cite{LFWTech}} 
& \multicolumn{2}{c|}{CALFW \cite{CALFW}} 
& \multicolumn{2}{c|}{CPLFW \cite{CPLFWTech}} 
& \multicolumn{2}{c|}{XQLFW \cite{XQLFW}} 
& \multicolumn{2}{c|}{Mean} \\

\cline{4-19}
 & & 
 & $1e{-3}$ & $1e{-4}$ 
 & $1e{-3}$ & $1e{-4}$ 
 & $1e{-3}$ & $1e{-4}$ 
 & $1e{-3}$ & $1e{-4}$ 
 & $1e{-3}$ & $1e{-4}$ 
 & $1e{-3}$ & $1e{-4}$ 
 & $1e{-3}$ & $1e{-4}$ 
 & $1e{-3}$ & $1e{-4}$ \\

\hline
\multirow{10}{*}{AUC-EDC}
& ViT-B @ 0-5   & 0-5 &
\textbf{0.0358} & \textbf{0.0808} & 0.0375 & 0.0524 & \textbf{0.0088} & \textbf{0.0131} & \textbf{0.0024} & \textbf{0.0030} & \textbf{0.0660} & \textbf{0.0724} & \textbf{0.0578} & \textbf{0.0786} & \textbf{0.2241} & \textbf{0.2714} & \textbf{0.0618} & \textbf{0.0817} \\
 & ViT-B @ 2-7                                   & 2-7 &
0.0749 & 0.1716 & 0.0355 & 0.0496 & 0.0293 & 0.0418 & 0.0045 & 0.0051 & 0.0835 & 0.0915 & 0.1434 & 0.1785 & 0.5502 & 0.6302 & 0.1316 & 0.1669 \\
 & ViT-B @ 4-9                                      & 4-9&
0.0655 & 0.1361 & 0.0295 & 0.0314 & 0.0335 & 0.0429 & 0.0023 & 0.0029 & 0.0670 & 0.0750 & 0.1287 & 0.1595 & 0.5536 & 0.6263 & 0.1257 & 0.1534 \\
  & ViT-B @ 6-11                                   & 6-11  &
0.0573 & 0.1318 & 0.0354 & 0.0412 & 0.0397 & 0.0489 & 0.0042 & 0.0050 & 0.0799 & 0.0865 & 0.1241 & 0.1633 & 0.4911 & 0.5467 & 0.1188 & 0.1462 \\
 & ViT-B @ 8-13                                      & 8-13&
0.0480 & 0.1108 & 0.0309 & 0.0452 & 0.0261 & 0.0352 & 0.0046 & 0.0054 & 0.0695 & 0.0762 & 0.1569 & 0.1989 & 0.5570 & 0.6110 & 0.1276 & 0.1547 \\
 & ViT-B @ 10-15                                     & 10-15&
0.0461 & 0.1040 & 0.0370 & 0.0516 & 0.0250 & 0.0350 & 0.0040 & 0.0050 & 0.0678 & 0.0748 & 0.1780 & 0.2235 & 0.5225 & 0.5666 & 0.1258 & 0.1515 \\
  & ViT-B @ 12-17                                     & 12-17&
0.0683 & 0.1578 & 0.0342 & 0.0405 & 0.0360 & 0.0424 & 0.0028 & 0.0035 & 0.0843 & 0.0927 & 0.1790 & 0.2213 & 0.4644 & 0.5843 & 0.1241 & 0.1632 \\
  & ViT-B @ 14-19                                & 14-19&
0.0654 & 0.1458 & 0.0335 & 0.0366 & 0.0397 & 0.0513 & 0.0036 & 0.0042 & 0.0773 & 0.0813 & 0.1306 & 0.1566 & 0.4717 & 0.5622 & 0.1174 & 0.1483 \\
 & ViT-B @ 16-21                                  & 16-21&
0.0574 & 0.1364 & 0.0294 & 0.0434 & 0.0264 & 0.0352 & 0.0044 & 0.0052 & 0.0802 & 0.0865 & 0.1694 & 0.2037 & 0.5539 & 0.6163 & 0.1316 & 0.1610 \\
 & ViT-B @ 18-23                                   & 18-23&
0.0640 & 0.1453 & \textbf{0.0258} & \textbf{0.0286} & 0.0363 & 0.0450 & 0.0038 & 0.0044 & 0.0702 & 0.0770 & 0.1229 & 0.1589 & 0.5070 & 0.5617 & 0.1186 & 0.1458 \\

\hline
\multirow{10}{*}{pAUC-EDC}
& ViT-B @ 0-5   & 0-5 &
\textbf{0.0127} & \textbf{0.0293} & 0.0090 & 0.0139 & \textbf{0.0048} & \textbf{0.0074} & \textbf{0.0007} & \textbf{0.0008} & \textbf{0.0187} & \textbf{0.0205} & \textbf{0.0240} & \textbf{0.0364} & \textbf{0.1256} & \textbf{0.1439} & \textbf{0.0279} & \textbf{0.0360} \\
 & ViT-B @ 2-7                                     & 2-7 &
0.0155 & 0.0355 & 0.0084 & 0.0134 & 0.0091 & 0.0138 & 0.0009 & 0.0011 & 0.0200 & 0.0223 & 0.0390 & 0.0527 & 0.1484 & 0.1665 & 0.0345 & 0.0436 \\
 & ViT-B @ 4-9                                     & 4-9&
0.0151 & 0.0342 & 0.0083 & \textbf{0.0096} & 0.0090 & 0.0127 & 0.0009 & 0.0010 & 0.0196 & 0.0219 & 0.0376 & 0.0490 & 0.1451 & 0.1693 & 0.0337 & 0.0425 \\
  & ViT-B @ 6-11                                    & 6-11  &
0.0145 & 0.0335 & 0.0088 & 0.0109 & 0.0092 & 0.0136 & 0.0009 & 0.0010 & 0.0204 & 0.0231 & 0.0353 & 0.0488 & 0.1489 & 0.1667 & 0.0340 & 0.0425 \\
 & ViT-B @ 8-13                                     & 8-13&
0.0147 & 0.0335 & \textbf{0.0082} & 0.0136 & 0.0083 & 0.0130 & 0.0009 & 0.0011 & 0.0196 & 0.0220 & 0.0430 & 0.0552 & 0.1493 & 0.1670 & 0.0349 & 0.0436 \\
 & ViT-B @ 10-15                                     & 10-15&
0.0140 & 0.0330 & 0.0095 & 0.0143 & 0.0079 & 0.0123 & 0.0010 & 0.0012 & 0.0197 & 0.0223 & 0.0466 & 0.0638 & 0.1495 & 0.1676 & 0.0355 & 0.0449 \\
  & ViT-B @ 12-17                                    & 12-17&
0.0154 & 0.0356 & 0.0086 & 0.0111 & 0.0095 & 0.0129 & \textbf{0.0007} & 0.0009 & 0.0205 & 0.0227 & 0.0491 & 0.0633 & 0.1435 & 0.1640 & 0.0353 & 0.0444 \\
  & ViT-B @ 14-19                                       & 14-19&
0.0146 & 0.0338 & 0.0084 & 0.0099 & 0.0098 & 0.0142 & 0.0008 & 0.0009 & 0.0200 & 0.0228 & 0.0363 & 0.0495 & 0.1383 & 0.1672 & 0.0326 & 0.0426 \\
 & ViT-B @ 16-21                                      & 16-21&
0.0152 & 0.0348 & 0.0089 & 0.0137 & 0.0085 & 0.0134 & 0.0009 & 0.0011 & 0.0199 & 0.0223 & 0.0436 & 0.0550 & 0.1485 & 0.1664 & 0.0351 & 0.0438 \\
 & ViT-B @ 18-23                                    & 18-23&
0.0150 & 0.0336 & 0.0085 & 0.0105 & 0.0090 & 0.0132 & 0.0008 & 0.0010 & 0.0196 & 0.0218 & 0.0371 & 0.0483 & 0.1408 & 0.1661 & 0.0330 & 0.0421 \\

\hline

\end{tabular}}
\end{table*}

\begin{table*}[ht!]
\centering
\setlength{\tabcolsep}{3pt}
\caption{Ablation studies analyzing four design choices: dataset generalization (WebFace4M, WebFace12M, CLIP, FRoundation), architecture depth (ViT-S vs ViT-B), block depth trade-offs (4-24 blocks), and aggregation strategies (uniform vs attention-weighted). Results show optimal performance at 12-20 blocks with last-block attention weighting. Mean AUC-EDC computed across seven benchmarks at FMR=$1e-3$ and $1e-4$. Best per study in bold.}
\label{tab:ablation_auc}
\vspace{-3mm}
\resizebox{\textwidth}{!}{
\begin{tabular}{|c|c|c|
cc|cc|cc|cc|cc|cc|cc|cc|}
\hline
\multirow{2}{*}{\textbf{Study}} & \multirow{2}{*}{\textbf{Method}}  & \multirow{2}{*}{\textbf{Blocks}}
& \multicolumn{2}{c|}{Adience \cite{Adience}} 
& \multicolumn{2}{c|}{AgeDB-30 \cite{agedb}} 
& \multicolumn{2}{c|}{CFP-FP \cite{cfp-fp}} 
& \multicolumn{2}{c|}{LFW \cite{LFWTech}} 
& \multicolumn{2}{c|}{CALFW \cite{CALFW}} 
& \multicolumn{2}{c|}{CPLFW \cite{CPLFWTech}} 
& \multicolumn{2}{c|}{XQLFW \cite{XQLFW}} 
& \multicolumn{2}{c|}{Mean AUC-EDC} \\

\cline{4-19}
 & &  
 & $1e{-3}$ & $1e{-4}$ 
 & $1e{-3}$ & $1e{-4}$ 
 & $1e{-3}$ & $1e{-4}$ 
 & $1e{-3}$ & $1e{-4}$ 
 & $1e{-3}$ & $1e{-4}$ 
 & $1e{-3}$ & $1e{-4}$ 
 & $1e{-3}$ & $1e{-4}$ 
 & $1e{-3}$ & $1e{-4}$ \\

\hline
\multirow{4}{*}{Dataset}
 & ViT-B - WebFace4M       & 0-23
 & \textbf{0.0217} & \textbf{0.0406} & 0.0255 & \textbf{0.0362} & 0.0133 & 0.0181 & \textbf{0.0034} & \textbf{0.0039} & 0.0707 & \textbf{0.0745} & 0.0538 & \textbf{0.0759} & \textbf{0.2575} & \textbf{0.2978} & \textbf{0.0637} & \textbf{0.0781} \\
 & ViT-B - WebFace12M      & 0-23
 & \textbf{0.0217}&	0.0417&	\textbf{0.0250}&	0.0393	&\textbf{0.0113}	&0\textbf{.0175}&	\textbf{0.0034}&	0.0040	&0.0907&	0.0965&	\textbf{0.0521}	&0.0805&	0.2994&	0.3391	&0.0719&	0.0884 \\
 & CLIP  & 0-11
 &  0.0634&	0.1368&	0.0364&	0.0508	&0.0351	&0.0491	&0.0033&	0.0040	&\textbf{0.0702}	&0.0772	&0.2049&	0.2542&	0.6065	&0.6645&	0.1457&	0.1767 \\
 & FRoundation   & 0-11 
 & 0.0605 &	0.1415	&0.0412	&0.0595&	0.0286	&0.0374	&0.0036	&0.0043	&0.0796	&0.0852&	0.1917&	0.2527	&0.5428&	0.5881&	0.1354	&0.1670 \\
\hline

\end{tabular}}
\end{table*}

\begin{table*}[ht!]
    \begin{center}
    \caption{The AUCs of EDC achieved by our method and the SOTA methods under different experimental settings. The notions of $1e-3$ and $1e-4$ indicate the value of the fixed FMR at which the EDC curves (FNMR vs.~reject) were calculated. The results are compared to three IQA and twelve FIQA approaches. The XQLFW dataset uses SER-FIQ (marked with *) as the FIQ labeling method.}
    \label{tab:sota_auc}
    \vspace{-3mm}
    \resizebox{0.99\textwidth}{!}{
\begin{tabular}{|c|cl|cc|cc|cc|cc|cc|cc|cc|cc|}
    \hline
\multirow{3}{*}{FR} & \multicolumn{2}{c|}{\multirow{3}{*}{Method}} 
& \multicolumn{2}{c|}{Adience} & \multicolumn{2}{c|}{AgeDB-30} & \multicolumn{2}{c|}{CFP-FP} 
& \multicolumn{2}{c|}{LFW} & \multicolumn{2}{c|}{CALFW} & \multicolumn{2}{c|}{CPLFW} 
& \multicolumn{2}{c|}{XQLFW} & \multicolumn{2}{c|}{IJB-C}\\ 
 & & 
 & \multicolumn{2}{c|}{\cite{Adience}} & \multicolumn{2}{c|}{\cite{agedb}} & \multicolumn{2}{c|}{\cite{cfp-fp}} 
 & \multicolumn{2}{c|}{\cite{LFWTech}} & \multicolumn{2}{c|}{\cite{CALFW}} & \multicolumn{2}{c|}{\cite{CPLFWTech}} 
 & \multicolumn{2}{c|}{\cite{XQLFW}} & \multicolumn{2}{c|}{\cite{ijbc}} \\ \cline{4-19}
                                                                                    &                                                   &  & $1e{-3}$ & $1e{-4}$ & $1e{-3}$ & $1e{-4}$  & $1e{-3}$ & $1e{-4}$  & $1e{-3}$ & $1e{-4}$  & $1e{-3}$ & $1e{-4}$ & $1e^{-3}$ & $1e{-4}$ & $1e{-3}$ & $1e{-4}$ & $1e{-3}$ & $1e{-4}$\\ \hline
     \hline 
     \multirow{16}{*}{\rotatebox[origin=c]{90}{ArcFace\cite{deng2019arcface}}}      & \multirow{3}{*}{\rotatebox[origin=c]{90}{IQA}}    & \multicolumn{1}{|l|}{BRISQUE\cite{BRISQE_IQA}} &0.0565 & 0.1285 &0.0400 & 0.0585 &0.0343 & 0.0433 &0.0043 & 0.0049 &0.0755 & 0.0813 &0.2558 & 0.3037 &0.6680 & 0.7122  & 0.0381	&0.0656                                                \\ 
                                                                                    &                                                   & \multicolumn{1}{|l|}{RankIQA\cite{liu2017rankiqa}} &0.0400 & 0.0933 &0.0372 & 0.0523 &0.0301 & 0.0384 &0.0039 & 0.0045 &0.0846 & 0.0915 &0.2437 & 0.2969 &0.6584 & 0.7039  & 0.0385	& 0.0640                                      \\
\hline 
    \end{tabular}}
    \end{center}
    \vspace{-5mm}
    \end{table*}

\begin{figure*}[t]
    \centering
    \setlength{\tabcolsep}{1pt}
    \renewcommand{\arraystretch}{1.1}
    \resizebox{0.9\textwidth}{!}{
        \includegraphics[width=0.44\textwidth]{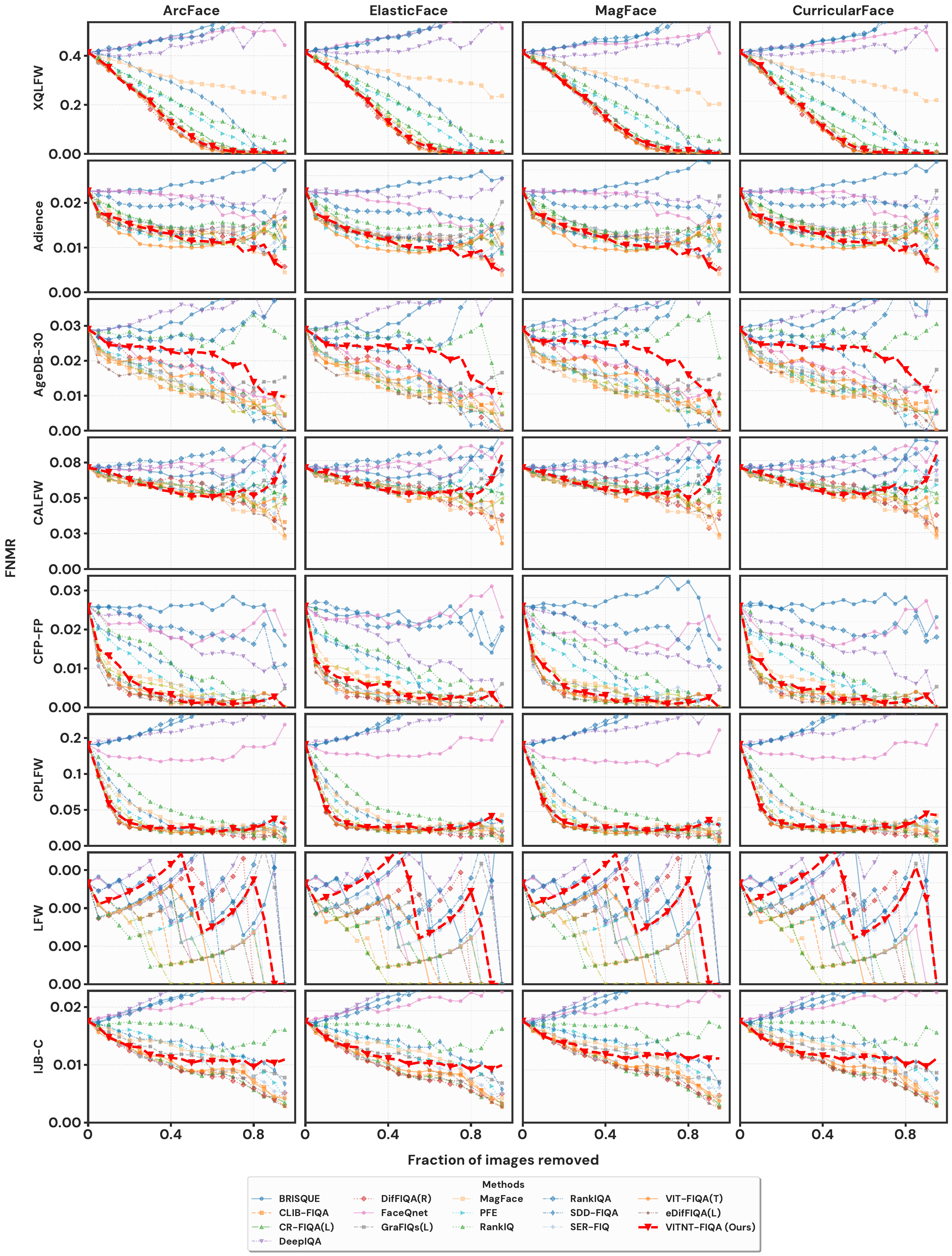}}
    \caption{Error-versus-Discard Characteristic (EDC) curves for FNMR@FMR=$1e-2$ of our proposed method in comparison to SOTA. Results shown on eight benchmark datasets: LFW \cite{LFWTech}, AgeDB-30 \cite{agedb}, CFP-FP \cite{cfp-fp}, CALFW \cite{CALFW}, Adience \cite{Adience}, CPLFW \cite{CPLFWTech}, XQLFW \cite{XQLFW}, and IJB-C \cite{ijbc}, using ArcFace \cite{deng2019arcface}, ElasticFace \cite{elasticface}, MagFace \cite{MagFace}, and CurricularFace \cite{curricularFace} FR models. Our method \ourmethod is marked with the red line.}
    \label{fig:fnmr2}
\end{figure*}

\begin{figure*}[t]
    \centering
    \setlength{\tabcolsep}{1pt}
    \renewcommand{\arraystretch}{1.1}
    \resizebox{0.9\textwidth}{!}{
        \includegraphics[width=0.44\textwidth]{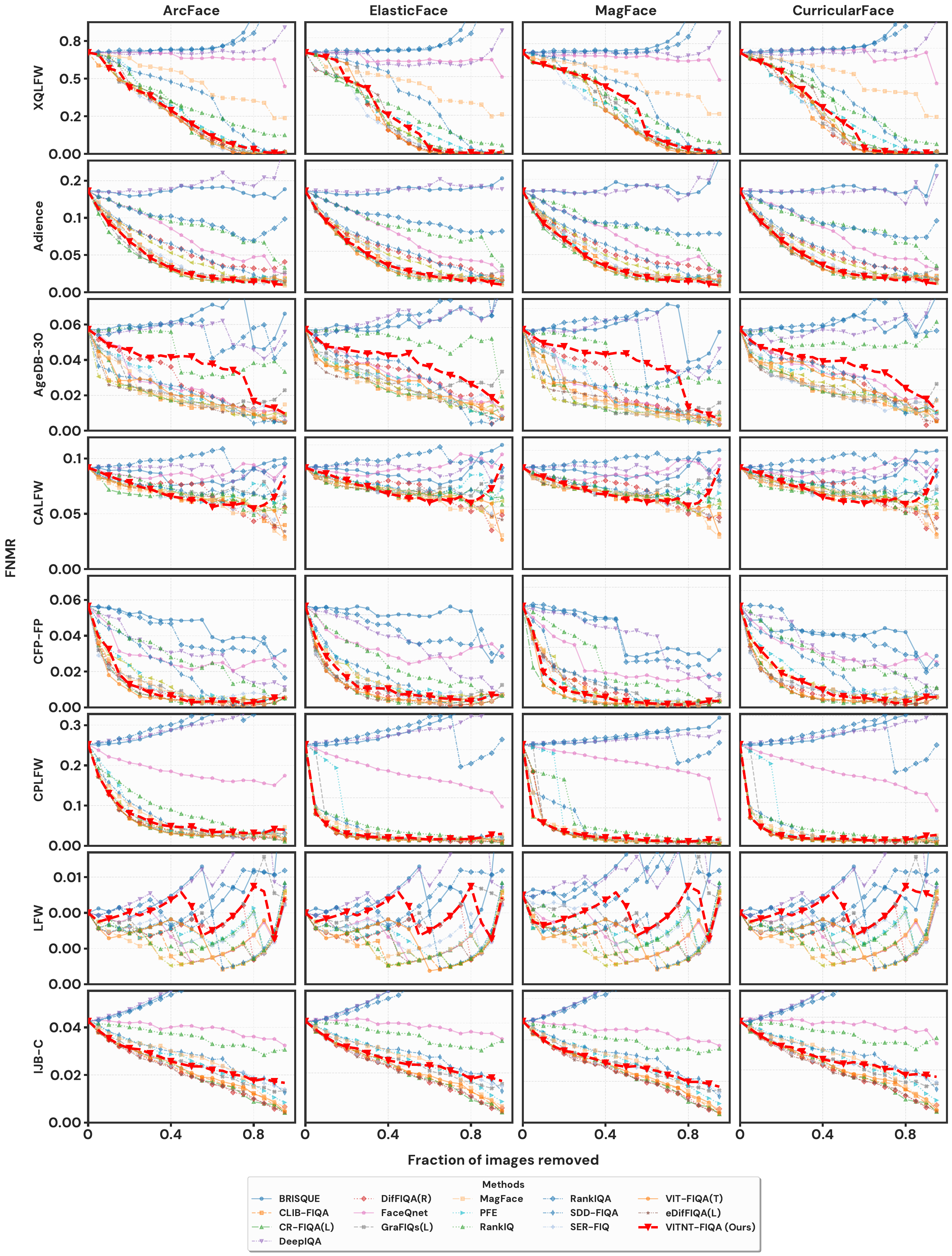}}
    \caption{Error-versus-Discard Characteristic (EDC) curves for FNMR@FMR=$1e-4$ of our proposed method in comparison to SOTA. Results shown on eight benchmark datasets: LFW \cite{LFWTech}, AgeDB-30 \cite{agedb}, CFP-FP \cite{cfp-fp}, CALFW \cite{CALFW}, Adience \cite{Adience}, CPLFW \cite{CPLFWTech}, XQLFW \cite{XQLFW}, and IJB-C \cite{ijbc}, using ArcFace \cite{deng2019arcface}, ElasticFace \cite{elasticface}, MagFace \cite{MagFace}, and CurricularFace \cite{curricularFace} FR models. Our method \ourmethod is marked with the red line.}
    \label{fig:fnmr4}
\end{figure*}

\begin{figure*}[t]
    \centering
    \setlength{\tabcolsep}{1pt}
    \renewcommand{\arraystretch}{1.1}
    \resizebox{\textwidth}{!}{
        \includegraphics[width=\textwidth]{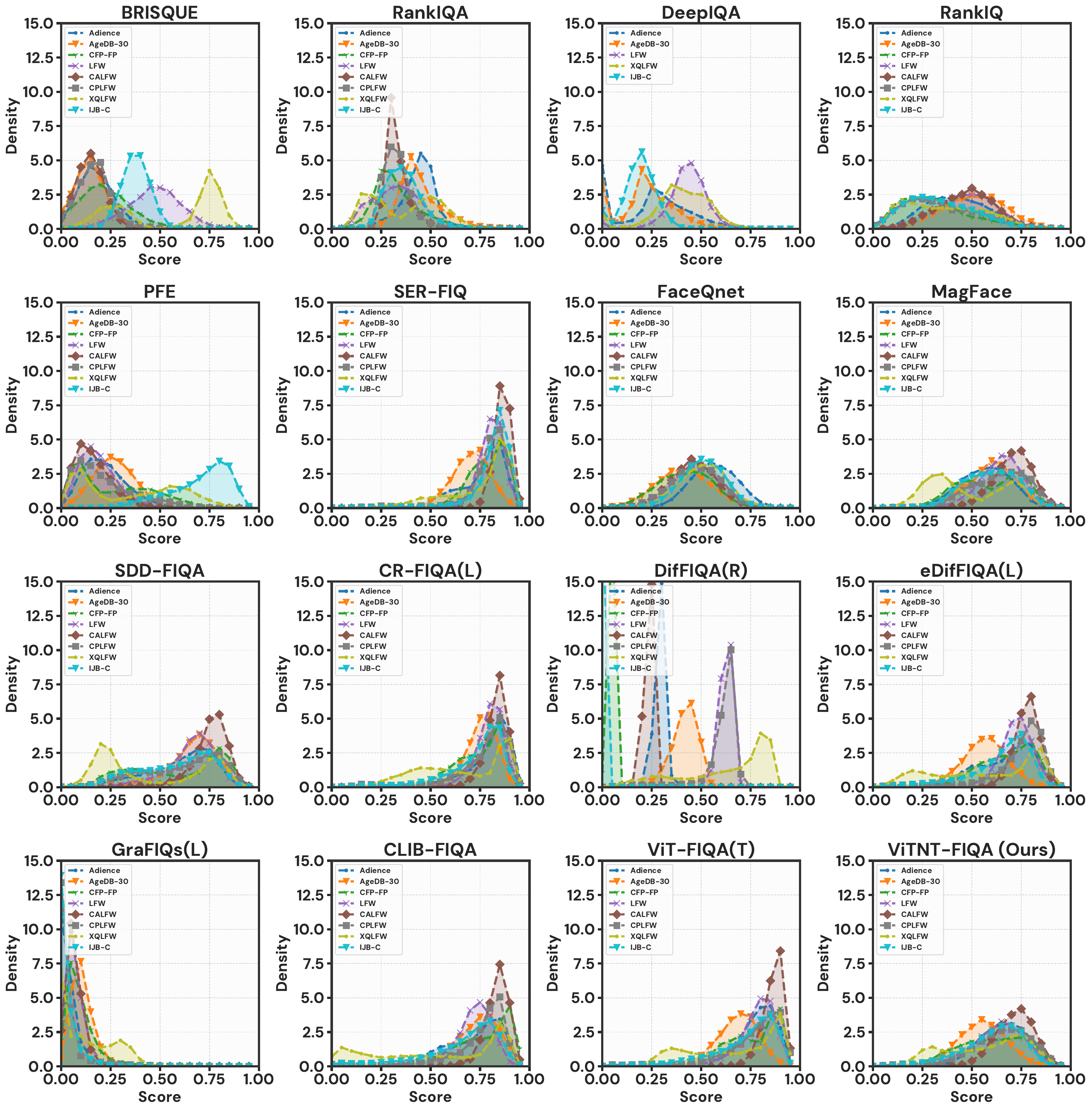}}
    \caption{Distribution of quality scores across the evaluation benchmarks, comparing our proposed method (\ourmethod) with SOTA methods. All scores are normalized to the range [0, 1].}
    \label{fig:quality_dist}
\end{figure*}

\end{document}